\documentclass{article}

    \usepackage[final]{neurips_2022}

\usepackage[utf8]{inputenc} 
\usepackage[T1]{fontenc}    
\usepackage{hyperref}       
\usepackage{url}            
\usepackage{booktabs}       
\usepackage{amsfonts}       
\usepackage{nicefrac}       
\usepackage{microtype}      

\usepackage{subcaption}
\usepackage{mwe}

\usepackage[dvipsnames]{xcolor}

\usepackage{amsmath}
\usepackage{mathtools}
\usepackage{graphicx}
\usepackage{algorithm}
\usepackage{algpseudocode}

\newcommand{\li}[1]{\textcolor{black}{#1}}
\newcommand{\pushi}[1]{\textcolor{black}{#1}}

\DeclareMathOperator*{\argmax}{arg\,max}

\title{
An Adaptive Deep RL Method for Non-Stationary Environments with Piecewise Stable Context
}

\author{%
  Xiaoyu Chen\thanks{Equal contribution. This work is conducted at Microsoft.}\textsuperscript{\hspace{0.6em}$\diamondsuit$} \\
  \And
  Xiangming Zhu\footnotemark[1]\textsuperscript{\hspace{0.6em}$\clubsuit$}\\
  \And
  Yufeng Zheng\textsuperscript{$\spadesuit$} \\
  \And
  Pushi Zhang\textsuperscript{$\ddag$} \\
  \AND
  Li Zhao\thanks{Corresponding author.}\textsuperscript{\hspace{0.6em}$\ddag$} \\
  \And
  Wenxue Cheng\textsuperscript{$\ddag$}\\
  \And
  Peng Cheng\textsuperscript{$\ddag$}\\
  \And
  Yongqiang Xiong\textsuperscript{$\ddag$} \\
  \AND
  Tao Qin\textsuperscript{$\ddag$} \\
  \And 
  Jianyu Chen\textsuperscript{$\diamondsuit$}\\
  \And
  Tie-Yan Liu\textsuperscript{$\ddag$} \\
  \And
  \centerline{$\diamondsuit$ Tsinghua University}\\
  \centerline{$\clubsuit$ Shanghai Jiao Tong University}\\
  \centerline{$\spadesuit$ University of California, Berkeley}\\
  \centerline{$\ddag$ Microsoft Research Asia}\\
}

\begin{document}

\maketitle

\vspace{-10pt}
\begin{abstract}

One of the key challenges in deploying RL to real-world applications is to adapt to variations of unknown environment contexts, such as changing terrains in robotic tasks and fluctuated bandwidth in congestion control. 
{
Existing works on adaptation to unknown environment contexts either assume the contexts are the same for the whole episode or assume the context variables are Markovian.}
However, in many real-world applications, the environment context usually stays stable for a stochastic period and then changes in an abrupt and unpredictable manner within an episode, resulting in a segment structure, which existing works fail to address.
To leverage the segment structure of piecewise stable context in real-world applications, in this paper, we propose a \textit{\textbf{Se}gmented \textbf{C}ontext \textbf{B}elief \textbf{A}ugmented \textbf{D}eep~(SeCBAD)} RL method. Our method can jointly infer the belief distribution over latent context with the posterior over segment length and perform more accurate belief context inference with observed data within the current context segment. The inferred belief context can be leveraged to augment the state, leading to a policy that can adapt to abrupt variations in context.
We demonstrate empirically that SeCBAD can infer context segment length accurately and outperform existing methods on a toy grid world environment and MuJoCo tasks with piecewise-stable context.
\vspace{-10pt}



\end{abstract}

\section{Introduction}
\label{Sec:1}
%

\vspace{-5pt}
Deep reinforcement learning has achieved great success in a wide range of challenging environments such as Atari games \citep{mnih2013atari,bellemare2012ale,hessel2017rainbow} or continuous control tasks \citep{schulman2015trpo, schulman2017ppo}. 
However, in stark contrast with this trend, applying RL to real-world applications remains a great challenge.
In most real-world settings, there could be variations in environmental factors, such as changing terrains in robotic tasks,  fluctuated bandwidth in congestion control, and dynamic traffic patterns in autonomous driving. We refer to such environment factors as \textit{situation} or \textit{context}. 
The changes in context are not neglectable since context usually has a substantial impact on transition and reward functions. 
When the context is fixed and known to us, the problem is stationary and easier to solve. However, in most realistic settings, the context is usually dynamic within an episode and unknown to us at test time.
Therefore, detecting and adapting to variations in context is very important for RL agents to make a real impact in a wide range of real-world applications. 

 \begin{figure}[t]
      \centering
      \includegraphics[width=0.8\textwidth]{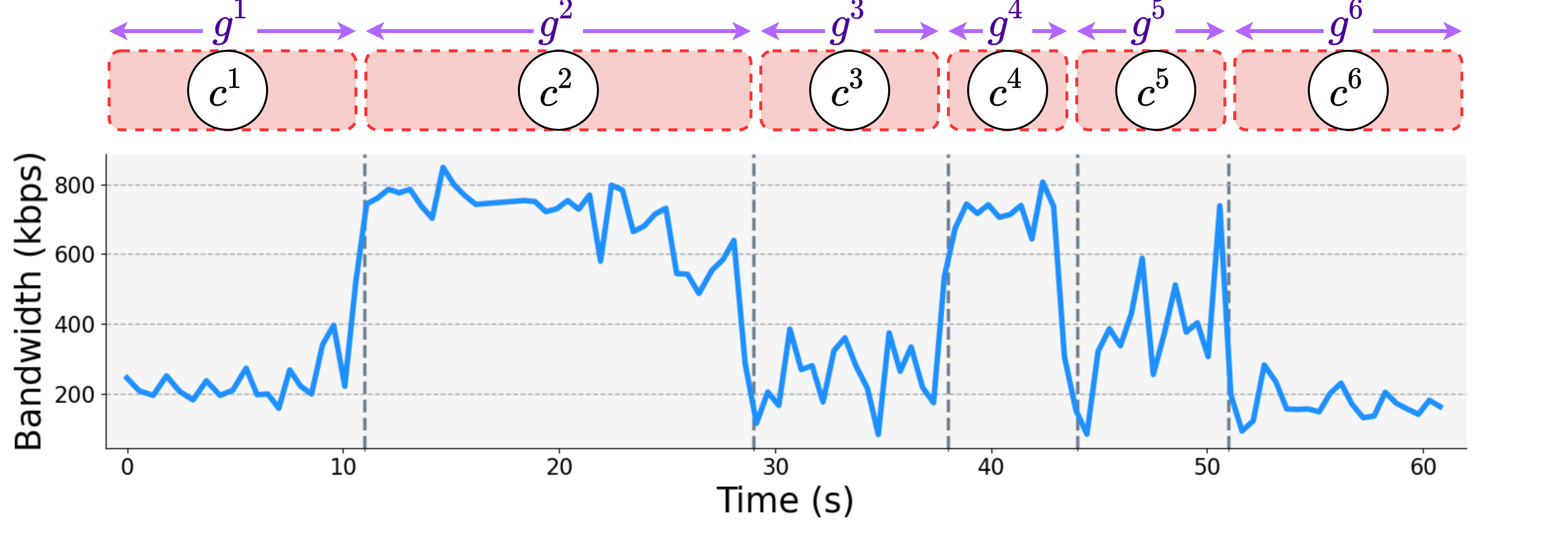}
      \caption{A typical trace of network available bandwidth. The trace can be approximately divided into several segments of different length $g^i$, for $i \in \{1,2,\cdots,6\}$. The network condition changes abruptly at the end of each \li{segment
      , while in each segment the network condition stays stable.
      }}
      \label{fluctuated_bandwidth}
\end{figure}

\vspace{-3pt}
In real-world environments with varied unknown contexts, we find a typical evolving context pattern particularly interesting. The context $c$ usually stays the same for a stochastic period until it changes abruptly and unpredictably into another context value $c'$ which is sampled i.i.d. from some prior context distribution. 
What's more, we usually do not have access to context $c$ directly but instead have access to some noisy observation $x_t$ sampled from some distribution $p(x_t|c)$ only at training time \li{(e.g., as auxiliary information from the simulator)}.
We take fluctuated bandwidth in congestion control as an example. 
In Figure \ref{fluctuated_bandwidth}, we show a typical trace of network available bandwidth. The fluctuated bandwidth is usually modeled by multiple non-overlapping segments \citep{akhtar2018oboe, zhang2001constancy}. Within each segment, the network condition is stationary, and thus the bandwidth approximately follows the same distribution.
At the end of a segment, the network condition changes, and the bandwidth changes abruptly into another distribution. Here the context $c$ represents the parameters that determine the distribution of fluctuated bandwidth (network condition), while the observations $x_t$ represents the observed bandwidth which follows the above distribution $p(x_t|c)$.
While the contexts $c$ are piecewise-stable, there could be slight variations in $x_t$ within each segment.
This piecewise-stable pattern of context dynamics is of particular interest to us since it can capture a wide range of stochastic context processes in real-world applications, such as changing terrains in robotic tasks and fluctuating bandwidth in congestion control. Since the change in context is abrupt and unpredictable, we cannot predict the future context in advance. The best we can do is detect and adapt to changes when they happen. 

\vspace{-3pt}
Although adaptation to varied unknown contexts has been studied under non-stationary RL and meta RL, few existing works look into piecewise stable context with abrupt changes within an episode.
Most works in meta RL as task inference~\citep{rakelly2019efficient,zintgraf2020varibad,zhao2020meld, poiani2021meta} and some works in non-stationary RL~\citep{chandak2020towards,chandak2020optimizing,xie2021deep} assume the context stays the same for the whole episode and infer the context based on the entire episode~(c.f. Figure \ref{fig:PGM}(a)), therefore cannot quickly adapt to context changes within an episode. 
Other works on non-stationary RL assume intra-episode context changes and model $c_t$ at each time step, but few study the piecewise-stable context as we do. 
\cite{nagabandi2018deep} directly predict the context and thus cannot capture the prior that the context tends to stay the same for a stochastic period.  
\cite{feng2022factored, ren2022reinforcement} model Markovian discrete context for each time step~(c.f. Figure \ref{fig:PGM}(b)), therefore failing to model the non-markov property of context and the prior over context segment length in our setting.
Compared with existing works, our setting is distinctive since the segment structure is latent to us
~(c.f. Figure \ref{fig:PGM}(c)).
Therefore, the unique challenge in our setting is that we need to infer the segment structure, which can be further leveraged to infer belief context by only incorporating the relevant observed data in the current segment. 

\vspace{-3pt}
\li{This paper studies how to infer segment structure to detect abrupt context changes and infer belief context accordingly to adapt to the piecewise-stable context in RL environments.}
We first introduce latent situational MDP which models RL environments with the stochastic \textit{situation}/\textit{context} process~(Section \ref{Sec:2}).  Then, we introduce how to infer the belief context from observed data. 
To address the challenge above, we propose to infer context segment structure and belief context jointly from observed data~(Section \ref{Sec:3-Inference}). 
Then we augment the state with inferred belief context so that the RL agent can automatically trade-off between acting optimally conditioned on inferred context and gathering more information about the current context~(Section \ref{Sec:3-policy}). 
Finally, we combine the training objectives for RL and inference and present all the details for our proposed deep RL algorithm~(Section \ref{Sec:3-practical}). 
We evaluate our algorithm on a gridworld environment with dynamic goals and MuJoCo tasks \citep{todorov2012mujoco} with varied contexts. Experiments demonstrate that our algorithm can quickly detect and adapt to abrupt changes in piece-wise stable contexts and outperform existing methods(Section \ref{experiment}).



\vspace{-4pt}
Our contributions can be listed as follows:
\begin{itemize}
\vspace{-6pt}
\item We introduce latent situational MDP with piecewise-stable context, which can capture a wide range of real-world applications~(Section \ref{Sec:2}). 
    \vspace{-2pt}
    \item We propose SeCBAD, an adaptive deep RL method for non-stationary environments with piecewise-stable context. Our method can infer the context segment structure and the belief context accordingly from observed data, which can be leveraged to detect and adapt to context changes~(Section \ref{method}). 
     \vspace{-2pt}
    \item Experiments on a gridworld environment and Mujuco tasks with piecewise-stable context demonstrate that our method can quickly detect and adapt to abrupt context changes and outperform existing methods~(Section \ref{experiment}).
\end{itemize}

\vspace{-5pt}
\section{Problem Formulation}
\label{Sec:2}

\begin{figure}
 \vspace{-10pt}
    \centering
    \includegraphics[width=0.99\linewidth]{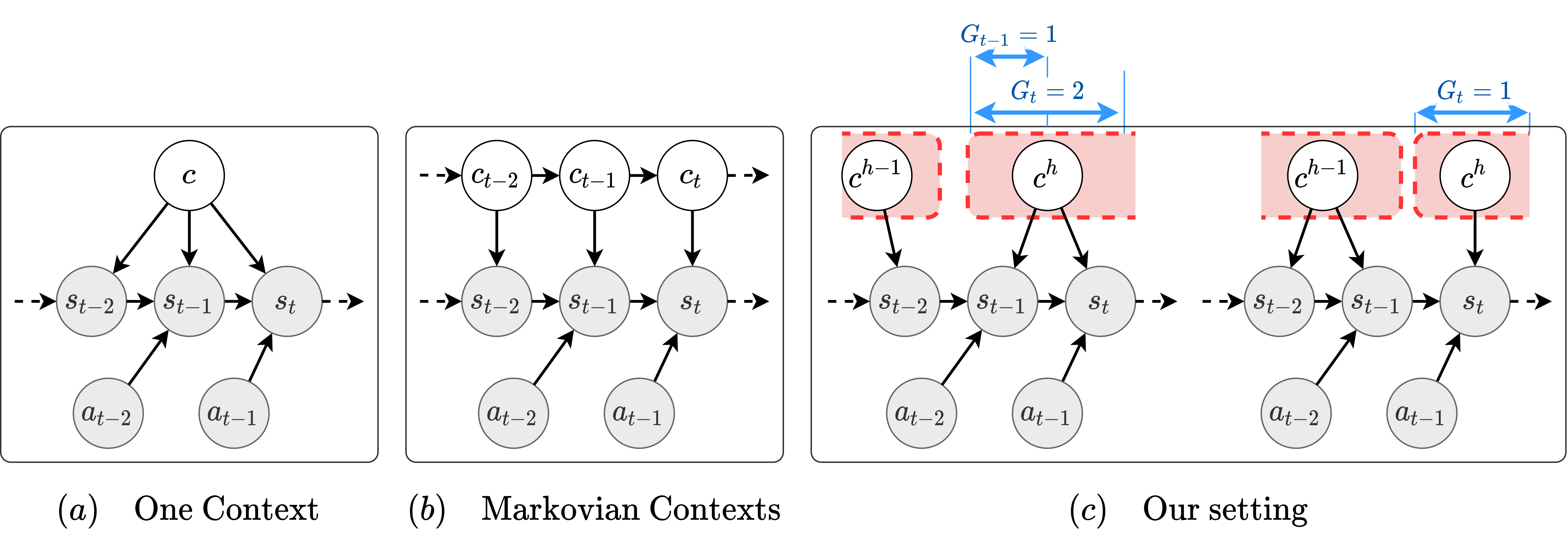}
    \caption{Probabilistic Graphical Models~(PGMs) for different problem settings, with shaded circles for observable variables and white circles for latent variables. (a) One context: assume the context remains unchanged for the whole episode. (b) Markovian context: assuming the context is Markovian for each time step. (c) Our setting: the context remains unchanged in each segment, but the segment structure, illustrated by the red segment is unknown and needs to be inferred. We show two possible examples of PGM corresponding to $G_t=2$~((c), left) and $G_t=1$~((c), right), where $G_t$ measures the length of the current segment up to time step $t$. } 
    \label{fig:PGM}
    \vspace{-5pt}
\end{figure}


\vspace{-5pt}
In this section, we define a \textit{latent situational Markov Decision Process (LS-MDP)} as a tuple $M=(\mathcal{S}, \mathcal{A}, \mathcal{C}, \mathcal{X}, G, T, R, \gamma)$, where $\mathcal{S}$ is the set of states, $\mathcal{A}$ is the set of actions, and $\gamma \in (0,1]$ is the discount factor.
To formulate the variations in environment factors, we introduce $\mathcal{C}$, the set of latent contexts, and $\mathcal{X}$, the set of observable contexts.
$\mathcal{C}$ refers to the set of underlying hidden contexts that contains all necessary information but are left unobservable to the agent, while $\mathcal{X}$ refers to the set of contexts with only partial information.
{{$G$ refers to the segment length which will be described in detail in the next paragraph.}}
$\mathcal{X}$ are observable only during training and unobservable during deployment. 
In an example environment setting where a robot walking over changing terrain, $\mathcal{C}$ represents the terrain features containing perfect information, and $\mathcal{X}$ represents noisy and imperfect information like mechanical metrics of the current step from virtual sensors in a simulator and thus is only accessible during training.

\vspace{-3pt}
In our setting, we focus on the case where the environmental changes are abrupt and irregular, \li{in contrast to the smoothly changing assumption on context/task in existing works in non-stationary RL~\citep{chandak2020towards,chandak2020optimizing}.} 
To better model the generative process of the contexts, we introduce the segment length $G$. 
Each episode is composed of
several stationary segments \li{with different segment lengths}.
For the ease of notation, we also introduce $G_t$, which measures the length of the current segment up to time step $t$.
Then, the generative process can be described as follows: 
at the beginning of the $h$-th segment, the environment samples $G^h \sim p_G(G)$ and $c^h \sim p_c(c)$ from 
\li{prior distributions $p_G$ and $p_c$.}
Then, in the next $G^h$ steps, the latent context $c^h$ remains unchanged. For each time step $t$ in this segment, the current segment length accumulates as $G_t=G_{t-1} + 1$ (we define $G_t = 1$ for the first time step in this segment), and the current latent context satisfies that $c_t = c^h$. The agent can observe $x_t \sim p(x_t | c_t)$ for each time step if during training. The stationarity lasts until the end of the current segment, then the environment resamples $G^{h+1}$ and $c^{h+1}$, and the process repeats. 
\li{We show two examples of graphical models of our setting given different $G_t$ in Figure \ref{fig:PGM}(c)}.

\vspace{-3pt}
The transition function $T: p(s_{t+1} | s_t, a_t, c^h)$ and the reward function $R: p(r_t | s_t, a_t, c^h)$ are all conditioned on current latent context $c^h$.
Therefore the changes in $\mathcal{C}$ lead to the changes in transition and reward functions. 
\li{At test time, $\mathcal{X}$ is no longer accessible, so we need to infer context changes from observed transitions and rewards. }
Since the latent context $\mathcal{C}$ remains unobservable and changes silently, the environment is no longer stationary for the agents. 
To act optimally, it is important to keep track of the environment to recognize changes in time, and rapidly adapt to those changes. 

\vspace{-10pt}
\section{Methodology}
\label{method}






\vspace{-5pt}
In this section, we present our \textit{Segmented Context Belief Augmented Deep~(SeCBAD)} RL method
and elaborate on how SeCBAD solves the challenges discussed above. Our method consists of two main components:
\begin{itemize}
    \vspace{-5pt}
    \item Joint inference of the belief distribution over the latent context and the segment structure from observed data.
    \item Policy optimization with inferred belief context under the belief MDP framework. 
\end{itemize}
\vspace{-5pt}
We first introduce the latent context inference part in Section \ref{Sec:3-Inference}, especially how to infer the segment structure jointly with belief context and leverage the segment structure to remove irrelevant observed data.
\li{After the belief context is approximated, it is then incorporated into the state as the input of the policy.} We detail the policy optimization part under the belief MDP framework
in Section \ref{Sec:3-policy}. 
And finally, in Section \ref{Sec:3-practical}, we describe how these parts constitute a practical algorithm.

\vspace{-5pt}
\subsection{Joint Inference for Belief Context and Segment Structure}
\label{Sec:3-Inference}


\vspace{-5pt}
In this part, we perform joint inference over latent context $c_t$ and current segment length $G_t$ from observed trajectory $\tau_{1:t}$, so as to remove irrelevant data in $\tau_{1:t}$ for belief context inference. This can be formally expressed by the following equation:
\begin{equation} \label{equation-1}
    p(c_t, G_t|\tau_{1:t}) = p(c^{t-G_t+1:t}|G_t, \tau_{t-G_t:t}) p(G_t|\tau_{1:t})
\end{equation}

where $\tau_{t_0:t} = (s_{t_0}, a_{t_0}, r_{t_0} \cdots, s_{t-1}, a_{t-1}, r_{t-1}, s_{t})$\footnote{This definition makes $\tau_{t-G_t:t}$ only contain observed data that belong to the current segment.}, and 
$c^{t-G_t+1:t}$ refers to the latent context for the whole segment from $t-G_t+1$ to $t$. In Equation \ref{equation-1}, we separately estimate the posterior $p(c^{t-G_t+1:t}|G_t, \tau_{t-G_t:t})$ of latent context given known segment structure $G_t$, and the posterior of segment length $p(G_t|\tau_{1:t})$.

\vspace{-5pt}
\subsubsection{Approximate Inference for Latent Context under Known Segment Structure}
\label{Sec:3.1-1inferknownGt}

\vspace{-5pt}
In this part, we focus on estimating $p(c^{t-G_t+1:t}|G_t, \tau_{t-G_t:t})$, which is the posterior of the latent context under known segment structure $G_t$.
We use the variational inference framework to approximate the true posterior. To be specific, we use an posterior inference network $q_\phi$ to infer the belief context in segment $[t-G_t+1:t]$: $q_\phi(c^{t-G_t+1:t}|G_t, \tau_{t-G_t:t})$.
The variational lower bound for the log-likelihood of the current segment is given by: 
\begin{align}
\label{Eq:elbo}
    &\quad \log p( \tau^X_{t-G_t:t} | a_{t-G_t:t-1}, s_{t-G_t}, G_t) \nonumber     \\
    &\geq  \mathbb{E}_{ q_\phi(c^{t-G_t+1:t}|G_t, \tau_{t-G_t:t})} \left[ \log p_{\theta}(\tau^X_{t-G_t:t}|G_t, c^{t-G_t+1:t}, a_{t-G_t:t-1}, s_{t-G_t}) \right] \nonumber \\
    &\quad\quad\quad\quad\quad\quad\quad\quad\quad
    -\mathbf{D}_{KL}(q_\phi (c^{t-G_t+1:t}|G_t, \tau_{t-G_t:t})\| p(c^{t-G_t+1:t})) \coloneqq \mathcal{J}_{Model}^t(G_t)
\end{align}


where $p_\theta$ denotes the decoder and $\tau^X_{t-G_t:t}= (\tau_{t-G_t:t}, x_{t-G_t+1:t})$.
For the detailed derivation, please see Appendix \ref{Appendix:ELBO}.

The reconstruction term of Equation \ref{Eq:elbo} can be factorized as:
\begin{align}
    &\quad \log p_{\theta}(\tau^X_{t-G_t:t}|G_t, c^{t-G_t+1:t}, a_{t-G_t:t-1}, s_{t-G_t}) \nonumber \\
    &= \sum_{i=t-G_t+1}^{t} \log p_{\theta}(x_{i}, s_{i}, r_{i-1}| G_t, c^{t-G_t+1:t}, s_{i-1}, a_{i-1})
\end{align}
which is the sum of log-probablity of transitions under context sampled from $q_\phi$. The term $\mathbf{D}_{KL}(q_\phi||p)$ is the KL-divergence between our variational posterior $q_\phi$ and the prior over the belief context.
For the prior $p(c^{t-G_t+1:t})$, we use previous posterior at timestep $t-1$ if the context remains unchanged at timestep $t$, or $\mathcal{N}(0, I)$ otherwise.



Unlike most previous methods which assume invariant context within an episode or markovian context at each time step, our method 1) takes only observed data within current segment $\tau_{t-G_t:t}$ as input for encoder $q_\phi$, 2) reconstructs only  data within current segment $\tau^X_{t-G_t:t}$ at the output for decoder $\log p_{\theta}$. Our method is naturally motivated by the piecewise-stable assumption on context dynamics. By removing irrelevant data outside the current segment \pushi{that are generated by past unassociated context}
, our method can estimate the current latent context more accurately. 



\begin{figure}[t]
    \centering
    \includegraphics[width=0.99\linewidth]{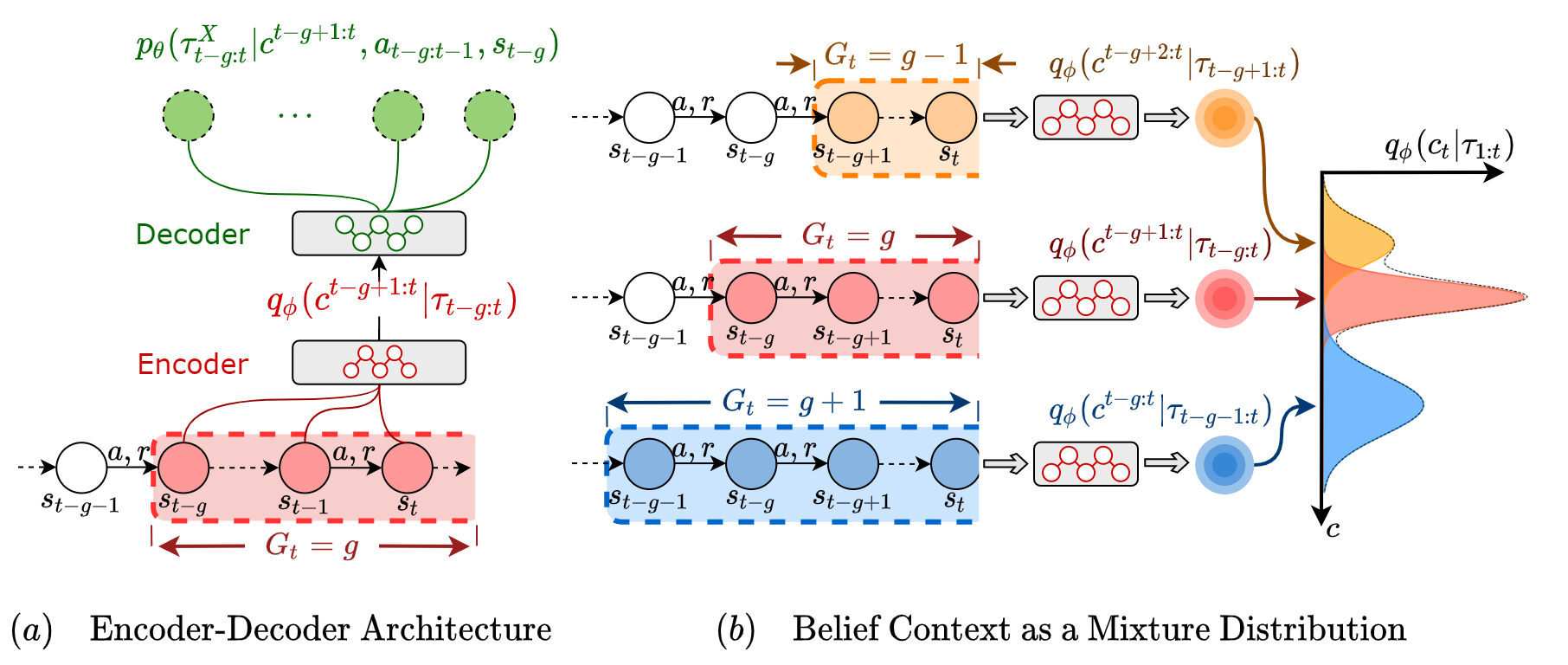}
    \caption{An overview of SeCBAD. 
    (a) Encoder-decoder architecture for belief context inference given current segment length $G_t = g$: the encoder $q_\phi$ takes the recent $g$ steps in the current red segment as input, and infer the belief context $q_\phi(c^{t-g+1:t}|\tau_{t-g:t})$, and the decoder takes in sampled context $c$ from $q_\phi$ and decode the trajectory segment $\tau^X_{t-g:t}$ (the green nodes). (b) Belief context $q_\phi(c_t|\tau_{1:t})$ as mixture distribution by considering belief context $q_\phi(c^{t-g+1:t}|\tau_{t-g:t})$ for all possible structures according to the posterior  $p(G_t|\tau_{1:t})$.}
    \label{Fig:algo}
    \vspace{-10pt}
\end{figure}

\subsubsection{Iterative Inference for the Segment Length}
\label{Sec:3recursive_inference}
In this part, we focus on estimating $p(G_t|\tau_{1:t})$, which is the posterior for current segment length, given $p(c^h|\tau_{t-G_t:t})$, the posterior of context under given segment structure.
To compute this posterior distribution, we first compute the joint distribution $p(G_t, \tau_{1:t})$ recursively based on $p(G_{t-1}, \tau_{1:t-1})$ as follows\footnote{For notation simplicity, we omit the condition $G_t=i$ in the term $p(s_t, a_{t-1}, r_{t-1}|\tau_{t-i:t-1})$, i.e. $t-i$ is the start of the segment.   
}:  
     \begin{align}
          \label{Eq:postG}
        p(G_t = i, \tau_{1:t}) &= \sum_{k=1}^{t-1} p(G_{t-1} = k, \tau_{1:t-1}) \cdot p(G_t = i| G_{t-1} = k) 
        \cdot p(s_t, a_{t-1}, r_{t-1}|\tau_{t-i:t-1})
     \end{align}

The three major components in Equation \ref{Eq:postG} in turn are the previous joint distribution, the evolution prior, and the observation probability. 
The previous joint distribution is iteratively provided at the beginning of each time step. The evolution prior $p(G_t=i|G_{t-1}=k)$ measures the prior knowledge on the segment length $G_t$ given the segment length of the previous time step $G_{t-1}=k$, where either $G_t=G_{t-1}+1$ or $G_t=1$ holds. 

As for the observation probability which is the third term of Equation \ref{Eq:postG}, it measures how likely the observations show up given the history of the segment. To be specific, we have \footnote{In Equation \ref{Eq:obs_prob}, $K = p(a_{t-1}|s_{t-1})$ is a constant with respect to $i$ and has no impact on the posterior $p(G_t=i|\tau_{1:t})$. 
}
\begin{align} \label{Eq:obs_prob}
    p(s_t, a_{t-1}, r_{t-1}|\tau_{t-i:t-1}) 
    &= K \mathbb{E}_{p(c^{t-i+1:t}|G_t=i, \tau_{t-i:t-1})} \Big[ p(s_t,r_{t-1}|s_{t-1}, a_{t-1}, c^{t-i+1:t}) \Big]
\end{align}


In Equation \ref{Eq:obs_prob}, the term $p(s_t,r_{t-1}|s_{t-1}, a_{t-1}, c^{t-i+1:t})$ estimates the data likelihood for the next state-reward pair, where $c$ is drawn from the posterior distribution given the data in the segment before timestep $t$.
To compute the RHS of Equation \ref{Eq:obs_prob}, we sample from $q_\phi(c^{t-i+1:t-1}|G_t=i, \tau_{t-i:t-1})$\footnote{When $i \geq 2$, we can use the belief $q_\phi(c^{t-i+1:t-1}|G_{t-1}=i-1, \tau_{t-i:t-1})$ to approximate the posterior $p(c^{t-i+1:t}|G_t=i, \tau_{t-i:t-1})$. For the case where $i=1$, the posterior distribution $p(c^{t-i+1:t}|G_t=i, \tau_{t-i:t-1})$ is actually the prior distribution $p_c(c)=\mathcal{N}(0,I)$, and we can sample $c$ from the prior distribution. } which is the belief context inferred from the whole history of the segment before timestep $t$, and use the sampled $c$ the decoder to compute the data likelihood.

Given the joint distribution, the posterior distribution of $G_t$ can be derived as 
\begin{equation}
\label{Eq:postG_norm}
    p(G_t = i | \tau_{1:t}) = \frac{p(G_t = i, \tau_{1:t})}{\sum_l p(G_t = l, \tau_{1:t})}
\end{equation}

We can also incorporate observable contexts $x$ in the observation probability to improve accuracy during training. See Appendix \ref{Appendix:obsx} for more details.



\vspace{-5pt}
\subsubsection{Belief Context as a Mixture Distribution}
\label{Sec:3.1-jointinference}



\vspace{-5pt}
Given inferred posterior of $G_t$ in Section \ref{Sec:3recursive_inference}, the belief of the latent context at time step $t$ can be derived as 
\begin{equation}
\label{Eq:q-marginalize}
    b_t(c) = q_\phi(c_t|\tau_{1:t}) = \sum_{g_t} q_\phi(c^{t-g_t+1:t}|G_t=g_t,\tau_{t-g_t:t}) p(G_t=g_t|\tau_{1:t})
\end{equation}


This mixed probability of $c^{t-g_t+1:t}$ which has taken all possible segment structures into consideration, can now represent the current belief $b_t$ of the latent context $c$. 

\vspace{-5pt}
\subsection{Policy Optimization with Belief Context}
\label{Sec:3-policy}

\vspace{-5pt}
Inspired by the belief MDP \citep{kaelbling1998planning}, Bayes Adaptive MDP (BAMDP) \citep{duff2002optimal} \li{and recent works on meta RL as task inference \citep{zintgraf2020varibad}, }
\li{we 
incorporate the inferred belief context into the augmented state.}
At \li{each} time step $t$, the belief latent context is approximated via $q_\phi$ using Equation \ref{Eq:q-marginalize}. 
Therefore, we define the augmented state as 
$(s, b) \in \mathcal{S} \times \mathcal{B}$, where $\mathcal{S}$ is the same state space as in \li{LS-MDP} and $\mathcal{B}$ is the set of belief latent contexts. \li{Accordingly, we have transition $T^{b}(s_{t+1}, r_t, b_{t+1} | s_t, a_t, b_t) = p(s_{t+1}, r_t|s_t, a_t, b_t) p(b_{t+1}|s_t, a_t, r_t, s_{t+1}, b_t)$ and reward $R^{b}(r_{t}|s_t, b_t, a_t)$ }
This definition brings advantages in the sense that the information gathering and exploitation tradeoff is no longer a problem under such augmented states, since the transition and reward functions are no longer conditioned on exact $c$.
Now, the policy is defined as $\pi(a|s, b)$ a mapping from the augmented state space to the action space, and the agent's objective is to maximize
\begin{align}
\label{Eq:RLloss}
\li{
    J_{RL} = \mathbb{E}_{s_0,b_0,\pi,T^{b}} \left[ \sum_{t=0}^{H} \gamma^t R^{b}(r_{t}|s_t, b_t, a_t)  \right] .}
\end{align}

\subsection{Algorithm and Implementations of SeCBAD}
\label{Sec:3-practical}




In this section, we describe the overall algorithm and implementation details of SeCBAD. 
See Figure \ref{Fig:algo} for an overview of our framework. As shown in Figure \ref{Fig:algo}(a), we use a GRU \citep{cho2014learning} parameterized by $\phi$ as the recurrent encoder $q_\phi$, and distributions in latent context space is assumed to be diagonal Gaussians with mean and variance parameterized by $q_\phi$. The decoder includes transition model $p_\theta(s_{t+1}|s_t, a_t, c)$, reward model $p_\theta(r_t|s_t, a_t, c)$ and observable context model $p_\theta(x_t|c)$. \li{The output of all the decoders are Gaussian distributions with mean parameterized by feed-forward neural networks and fixed identity covariance.} Then, we estimate $p(G_t|\tau_{1:t})$ using $q_\phi$ and $p_\theta$ as described in Section \ref{Sec:3recursive_inference}.
As shown in Figure \ref{Fig:algo}(b), we combine the belief context based on different segment according to $p(G_t|\tau_{1:t})$ to get the belief $b_t(c) = q_\phi (c_t|\tau_{1:t})$, where one approach is to provide a total of $t$ mean, covariance and weights as policy input. For simplicity, we choose $G_t^*$ with highest probability in $p(G_t|\tau_{1:t})$ and use the corresponding $q_\phi(c^{t-G_t^*+1:t}|\tau_{t-G_t^*:t})$ as the belief. Empirically, we find this approximation  leads to little performance loss.
We build our RL algorithm on the top of PPO \citep{schulman2017proximal} to learn the policy $\pi_{\psi}(a_t|s_t,b_t(c))$, where $\psi$ denotes the parameters in the actor and the critic. We use the objective described in Section \ref{Sec:3-policy} to optimize the policy.

During deployment, $q_\phi$ and $p_\theta$ are fixed. We first compute the segment posterior $p(G_t|\tau_{1:t})$ using encoder $q_\phi$ and the transition and reward model $p_\theta$. Then, we estimate the belief $b_t(c)$ and feed it into the policy as input. 
For more implementation details, please refer to the Appendix \ref{Appendix:implementation}.

\section{Experiments}
\label{experiment}

{{
In this section, we empirically evaluate 
SeCBAD
on two tasks. We first demonstrate our algorithm in a grid world environment in Section \ref{Sec:4-1gridworld} to illustrate the significance of incorporating segment structure during inference.
For large-scale experiments, we test the proposed algorithm in Section \ref{Sec:4-2mujoco}.     The results of multiple challenging tasks show that SeCBAD outperforms baselines in terms of performance and sample efficiency.
In Section \ref{Sec:4-3ablation}, we further provide case studies of agent behaviors and learned latent to gain more insights into the results. 
}}




\subsection{Motivating Example}
\label{Sec:4-1gridworld}

\begin{figure*}[t]
        \centering
        \begin{subfigure}[b]{0.7\textwidth}  
            \centering 
            \includegraphics[width=0.8\textwidth]{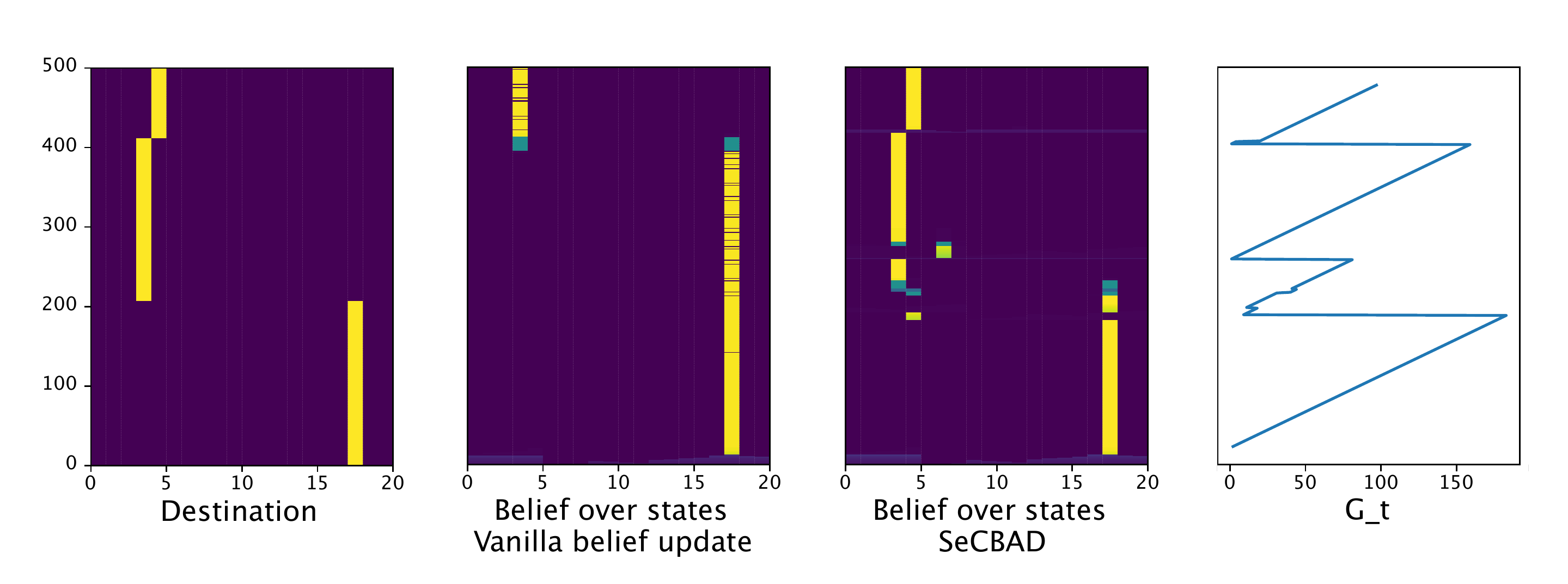}
            \label{fig:exp_4_1-2}
        \end{subfigure}
        \caption[]
        {Experiment on grid world with dynamic goal state. 
        We show a case study on the inferred belief. The leftmost figure depicts the ground truth goal position, and the next two figure depicts the belief estimated by vanilla inference and SeCBAD. where the color of one grid measures how the agent believes $s^*$ locates at that grid. The rightmost figure is $G_t^*$ estimated by SeCBAD. The result shows that our belief matches the ground truth more closely.
        }
        \label{fig:exp_4_1}
        \vspace{-15pt}
    \end{figure*}

\vspace{-5pt}
In this experiment, we use a motivating example to show that it is vital for the agent to consider the segment structure to adapt to variations of unknown environment contexts. The agent is in a $4 \times 5$ grid world containing 1 goal state $s^*$ and 19 other states. In $s^*$, the agent can gain $+1$ reward with probability $p_0$, and $0$ reward with probability $1 - p_0$. For other states, the agent can gain $+1$ reward with probability $p_1$ and $0$ reward with probability $1 - p_1$, where $p_0 > p_1$. The actions include up, down, left, right, and none. The specific location of $s^*$ is unobservable to the agent, and may change after a stochastic period.

\vspace{-3pt}
We use the analytical belief update rule instead of variational inference in this experiment. To be specific, we assume the $q_\phi$ and $p_\theta$ is known and fixed and then infer $p(G_t|\tau_{1:t})$ and the belief $b_t(c)$, which represents the belief distribution of the location of $s^*$. The belief update rule is detailed in Appendix \ref{app:bayes_update}. With the analytical belief update rule, a theoretically optimal belief $b_t$ is computed for each environment step $t$. We concatenate $(s_t, b_t)$ together as the augmented state. We may expect that a policy trained upon this augmented state will achieve better performance if $b_t$ is an accurate enough guess to the true location of the goal state. In such a way, we aim to focus on the effectiveness of inferring segment structure, in terms of how inferring segment structure can arrive at a meaningful belief state. We compare our algorithm with vanilla inference, where the agent updates belief without considering segment structure, like in \cite{zintgraf2020varibad, rakelly2019efficient}. For both of two routines of updating belief state, we let the agent move along a same path. By moving along a same path, we mean the agent starts from the same location and then take the same action in each environment step for two belief update rules.

\vspace{-3pt}
Figure \ref{fig:exp_4_1} illustrates how SeCBAD and the vanilla inference baseline update their belief over states. It is shown that the belief $b_t(c)$ of SeCBAD closely matches with the actual goal state $s^*$. Also, compared with SeCBAD, the vanilla inference baseline needs more steps to detect the changes since it needs to correct the deviated prior belief. This result supports that incorporating segment structure into inference like SeCBAD is significant for estimating accurate belief in fast varying environments.

\vspace{-8pt}
\subsection{Locomotion Control Tasks with Varying Contexts}
\label{Sec:4-2mujoco}

\begin{figure*}[t]
    \centering
    \begin{subfigure}[b]{0.99\textwidth}
        \centering
        \includegraphics[width=\textwidth]{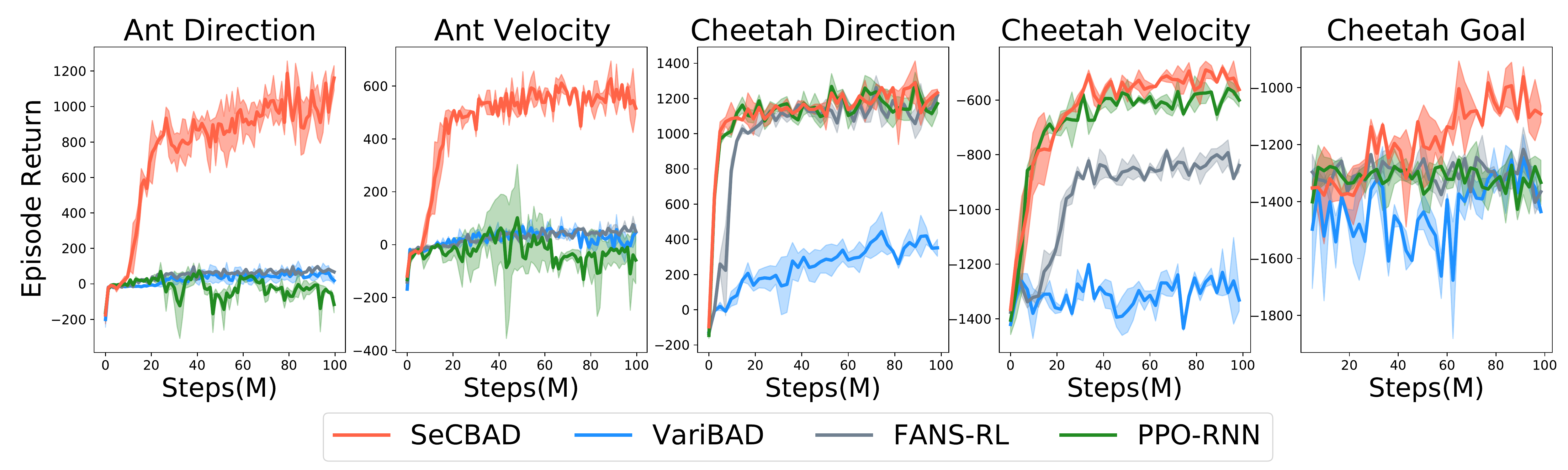}
    \end{subfigure}
    \caption[]
        {{Performance curves on 5 MuJoCo environments with variations in contexts. SeCBAD achieves better performance and sample efficiency in various challenging control tasks.}}
        \label{fig:exp_4_2_reward_curve}
        \vspace{-10pt}
\end{figure*}

\vspace{-5pt}
{{
In this section, we show that SecBAD is able to tackle more complex tasks by testing the algorithm on several challenging control tasks with varying contexts.
We conduct the experiments on modified MuJoCo \citep{todorov2012mujoco} tasks.
These environments are commonly used in previous works \citep{zintgraf2020varibad, rakelly2019efficient}, and we further modify the contexts to provide challenges corresponding to LS-MDP as mentioned in Section \ref{Sec:2}. 
As for Ant Direction and Cheetah Direction, the policy needs to move towards the target direction as fast as possible. 
As for Ant Velocity and Cheetah Velocity, the policy then needs to additionally move at the speed of the target velocity besides the target direction. 
For Cheetah Goal, the policy needs to navigate to varying goals and stay there.
For Ant environments, the direction and velocity are specified on a 2-dimensional plane, and for half cheetah environments, the direction and velocity are 1-dimensional.
In these environments, the contexts refer to the target velocity, target direction, or location of the goal. To make it more challenging, we further add noises to contexts within each segment. Please refer to Appendix \ref{Appendix:implementation} for more details.
}}
{{As for baselines, we select one representative algorithm for each type in Figure \ref{fig:PGM} to compare since to the best of our knowledge, few existing works study the same setting as we do. 
We use VariBAD \citep{zintgraf2020varibad} to represent those methods assuming one context in the episode, and use FANS-RL \citep{feng2022factored} to represent those methods assuming Markovian contexts. Since LS-MDP is a special case of POMDP, we include a POMDP baseline PPO-RNN \citep{hausknecht2015deep} for comparison.}}
{\textcolor{red}{
}}
{{
The results are provided in Figure \ref{fig:exp_4_2_reward_curve}. 
}}

\vspace{-3pt}
{{
The experiment results empirically show that SeCBAD achieves superior performance and sample efficiency on challenging control tasks. 
As illustrated in Figure \ref{fig:exp_4_2_reward_curve}, SeCBAD achieves higher scores than baselines.
We provide some insights into the results. For the method assuming that contexts stay the same within an episode, \citep{zintgraf2020varibad} uses the learned latent contexts to decode the whole trajectory including transitions and rewards in other segments. This reconstruction mismatch may lead to averaged latent contexts so that the policy cannot act correspondingly
(see detailed analysis and the ablation study in Appendix \ref{ablation:numberofsegment}.)
}}
{{
For Markovian contexts baseline \citep{feng2022factored} and POMDP baseline \citep{hausknecht2015deep}, they can perform relatively well on tasks where the contexts can be inferred from only one step transition (i.e., Cheetah Direction and Cheetah Velocity).
However, for more complex tasks where the contexts need more steps to infer, SeCBAD significantly outperforms these two baselines \citep{feng2022factored, hausknecht2015deep}, which proves that the specially designed joint inference component in SeCBAD is effective and can help improve the performance.
}}

{{
To better understand the proposed algorithm, we conduct a series of ablation studies. In Appendix \ref{app:prior}, we study the effects of inaccurate prior on SeCBAD. In Appendix \ref{ablation:pgt}, we study the implementation choices on how to use $p(G_t|\tau_{1:t})$. In Appendix \ref{ablation:robustness}, we further test SeCBAD against different levels of noises within each segment.
}}

\vspace{-5pt}
\subsection{Case Studies on the Segment Structure}
\label{Sec:4-3ablation}

\begin{figure*}[t]
    \centering
    \begin{subfigure}[t]{0.6\textwidth}
        \centering
        \includegraphics[width=\textwidth]{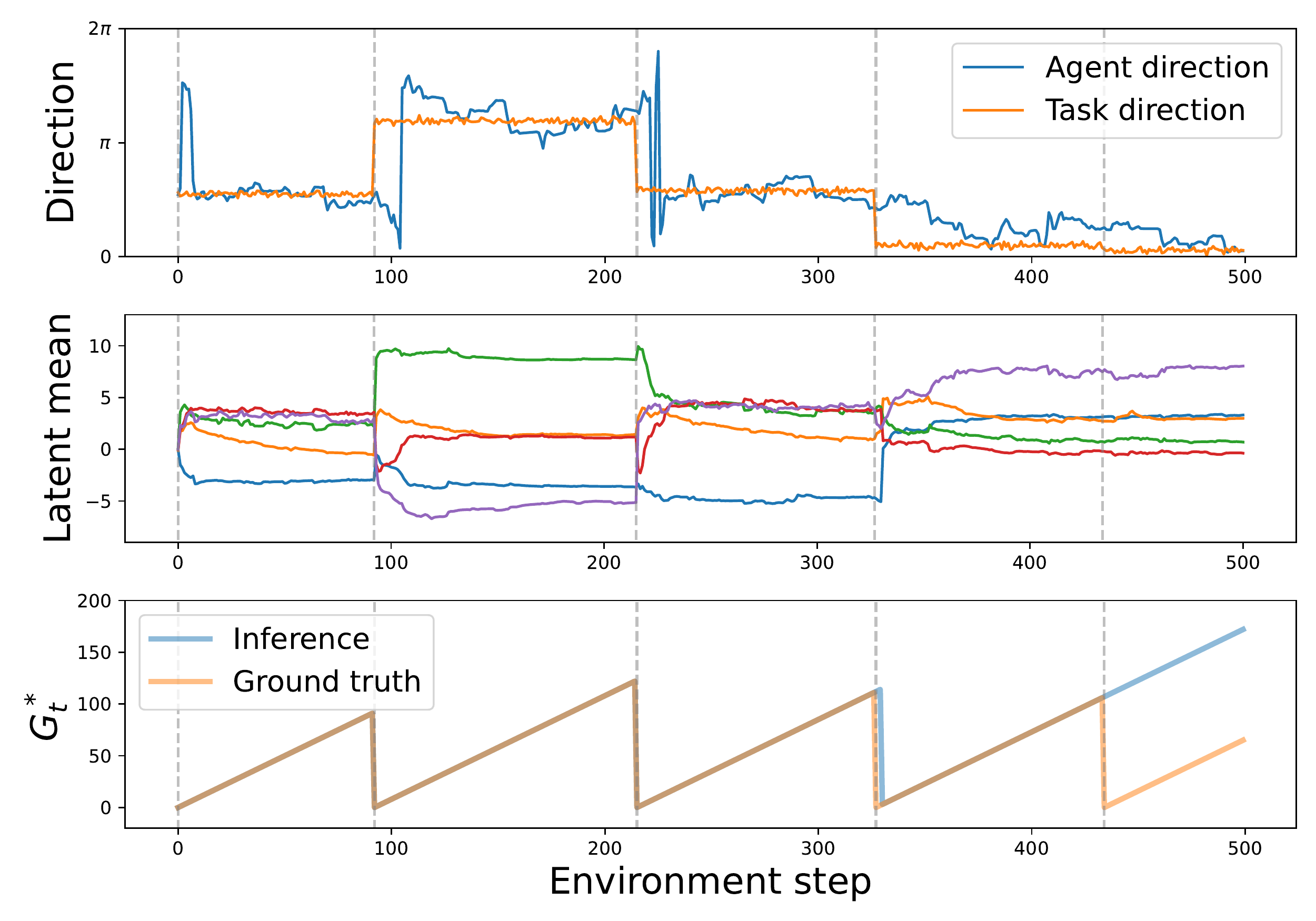}
        \label{fig:exp_4_3_trunc_point}
    \end{subfigure}
    \caption[]
        {{A case study on Ant Direction of SeCBAD. In the first row, we show the behavior of the agent with the goal direction in orange and the actual agent direction in blue. In the second row, we show the learned latent contexts with different colors corresponding to different dimensions. In the last row, we plot the $G_t^*$  inferred by SeCBAD.}}
        \label{fig:exp_4_3}
        \vspace{-15pt}
\end{figure*}

\vspace{-3pt}
{{
In order to provide more insights into the results, we present a case study on Ant Direction in this section. We show the behavior and learned latent along with the inferred segment of a randomly selected episode during testing of the SeCBAD algorithm on Ant Direction in Figure \ref{fig:exp_4_3}.
}}

{
In the first row, we exhibit the task direction in orange as well as the agent's actual direction in blue. It shows that SeCBAD can rapidly detect and adapt to context changes.
We also show the learned latent contexts in the second row and the inferred segment structure in the third row. From the third row, it can be seen that the agent can detect the segment in time as the inferred segment structure closely matches the ground truth segment structure.  As for the segment starting from the 430th environment step, the change in goal direction is negligible so that the agent automatically merges the two segments.
Since the latent contexts are calculated according to the inferred $G_t^*$, the latent contexts in the second-row change as the segment changes while staying stable within each segment. This proves that the agent can be aware of the changes in unknown contexts and quickly adapt to the changes. We provide case studies of baselines and detailed analysis in Appendix \ref{app:casestudy}.
}

\subsection{Bandwith Control Tasks for Real Time Communication}

To further illustrate that the proposed LS-MDP setting can boost the deployment of RL in many real-world applications, we test SeCBAD on a real-world bandwidth control task for real-time communications (RTC) \citep{alphartc} in this section.
The most critical goal in RTC is to provide high Quality of Experience (QoE) for users.
To achieve this, a bandwidth control module is needed, i.e., the RTC sender needs to decide the bitrate of outstreaming audio/video based on the network status towards the receiver. 
However, the network conditions constantly change and the changes occur in multiple items, which makes this problem intricate and hard to solve.

In our reinforcement learning formulation, we use a 7-tuple of current network statistics that is visible to the agent as states $s_t$, consisting of sending rate, short-term and long-term receiving rate, loss, and delay.
The action $a_t$ is the estimated bandwidth. 
We provide the detailed settings in Appendix \ref{app:rtcexp}.
The latent context $c_t$ here refers to the fluctuated network condition, in this section, we consider $x_t$ as the ground truth bandwidth capacity $C$, and the RTT.
{
In this experiment, we compare our method with VariBAD \citep{zintgraf2020varibad}, FANS-RL \citep{feng2022factored}, vanilla PPO \citep{schulman2017ppo} and PPO-RNN \citep{hausknecht2015deep}. 
We also incorporate oracle PPO scores by incorporating the unobservable contexts into the observable states}. {{All the methods are trained for 10 million steps and the shaded area is across 3 random seeds.}}

\begin{figure}[t]
    \centering
    \includegraphics[width=0.4\textwidth]{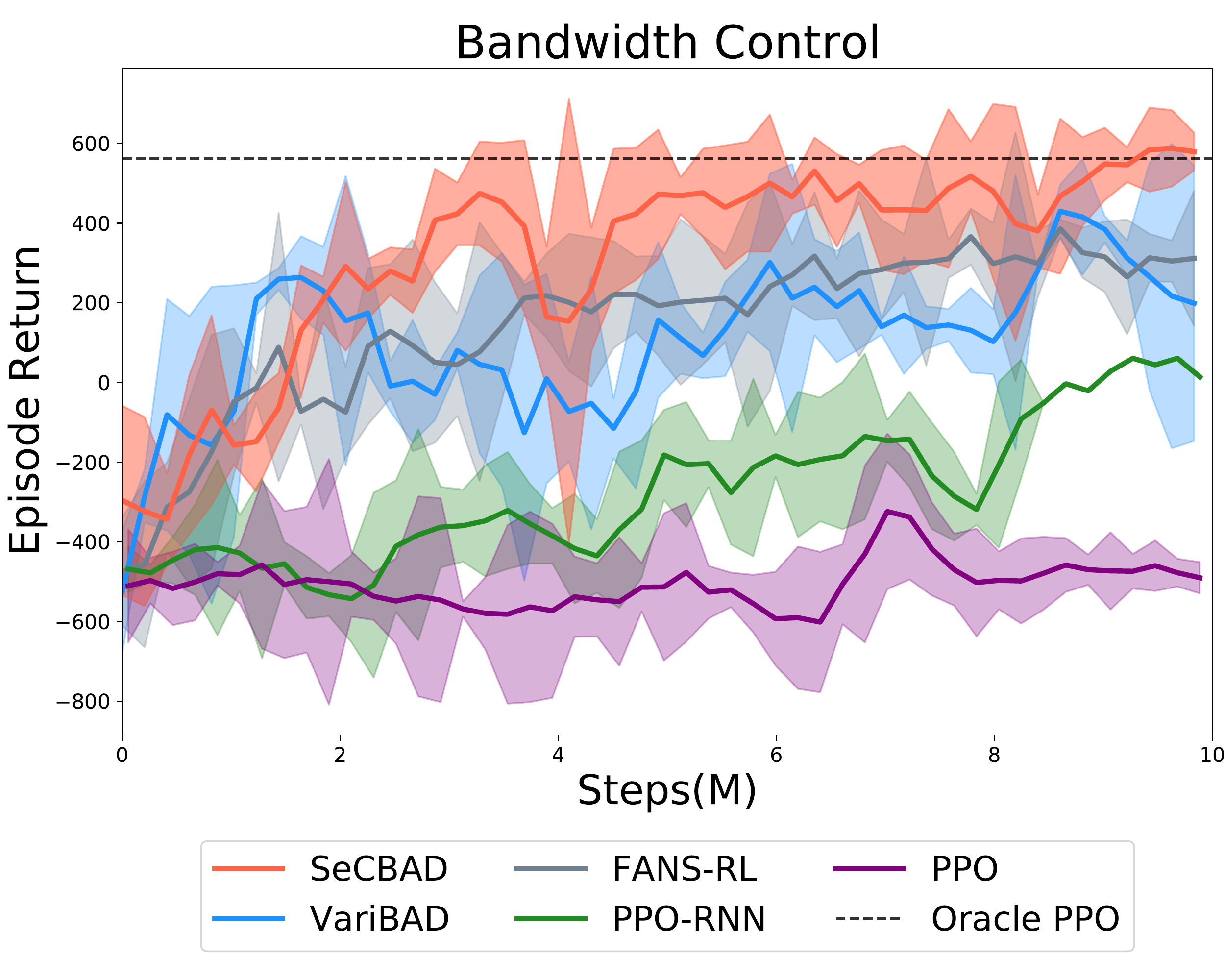}
    \caption{Experiment results on bandwidth control for RTC.}
    \label{fig:exp-rtc-main}
\end{figure}

{As illustrated in Figure \ref{fig:exp-rtc-main}, SeCBAD achieves better performance than other baselines and is very close to the oracle PPO baseline score. The performances of VariBAD \citep{zintgraf2020varibad} and FANS-RL \citep{feng2022factored} are better than PPO-RNN \citep{hausknecht2015deep}, but SeCBAD outperforms both of these methods. The results suggest that SeCBAD is able to detect and adapt to the varying contexts more rapidly, which allows the policy to precisely control. 
For detailed descriptions and results, please refer to Appendix \ref{app:rtcexp}.
}

\vspace{-5pt}
\section{Related works}

Our work is closely related to \textbf{non-stationary RL}, where the transition and reward functions may change over time. Most existing works on non-stationary RL focus on inter-episode non-stationarity~\citep{xie2021deep, chandak2020towards, chandak2020optimizing, xie2022robust, sodhani2021block, alegre2021minimum, poiani2021meta, al2017continuous}, some of which adopt contextual MDP as formulation \citep{hallak2015contextual}. Recently there are some works considering intra-episode non-stationarity~\citep{ren2022reinforcement, kamienny2020learning, kumar2021rma, nagabandi2018deep, feng2022factored}.
\cite{ren2022reinforcement} assume the latent context to be finite and Markovian, while \cite{feng2022factored} assume the latent context is Makovian and the environment can be modeled as a factored MDP. 
In contrast to existing works on non-stationary MDP, we assume piecewise-stable context with abrupt changes within an episode, which is more realistic and can capture a wide range of real-world applications. To rapidly adapt to dynamic context, the agent needs to continuously perform information gathering behavior.
\textbf{Bayesian RL} \citep{duff2002optimal, zintgraf2020varibad, fellows2021bayesian} is an elegant framework to optimally tradeoff the exploration and exploitation in an unknown and stationary MDP. As a special type of \textbf{Belief MDP} \citep{kaelbling1998planning}, Bayes Adaptive MDP (BAMDP) \citep{duff2002optimal} maintains a belief over the environment and uses this belief to augment the state.
\li{Our model can be viewed as a special case of belief MDP, where we only maintain belief over latent context to trade off between information gathering and exploitation. 
}
To accurately infer belief context, we adopt the \textbf{variational inference}, 
which has been adopted by many existing works in RL for task inference \citep{rakelly2019efficient, zhao2020meld, humplik2019meta, poiani2021meta, zintgraf2020varibad} or context inference \citep{xie2021deep, feng2022factored, ren2022reinforcement}.
However, none of these methods suit our setting. \li{We perform joint inference over latent context and segment structure from observed data, so as to remove irrelevant data for more accurate belief context inference. }
LS-MDP can also be viewed as a special case of \textbf{POMDP}. Recently, progress has been made in learning the latent dynamics model \citep{krishnan2015deep,karl2016deep,doerr2018probabilistic,buesing2018learning,ha2018world,han2019variational,hafner2019learning,hafner2019dream}. 
Theoretically, it is possible to perform optimally only using recurrent neural networks (RNNs) like \cite{hausknecht2015deep} since the whole history has been taken into consideration.
However, it has been shown that \citep{hafner2019learning} introducing more structured information will significantly enhance the performance. In this paper, we 
\li{exploit the addtional assumption of LS-MDP} to infer the context based on segment structure. 
The experiments show that SeCBAD can achieve better performance compared with \li{existing methods}. 

\vspace{-10pt}
\section{Conclusion}

\vspace{-8pt}
In this paper, we propose SeCBAD, a \textbf{Se}gmented \textbf{C}ontext \textbf{B}elief \textbf{A}ugmented \textbf{D}eep RL method to deal with piecewise-stable context in non-stationary environments. Piecewise-stable context is quite common in a wide range of real-world applications. Compared with existing methods, our method can automatically detect the segment structure, which reflects when the context changes abruptly. The detected segment structure can be further used to compute context belief with only relevant observed data. To the best of our knowledge, this is the first method that can model and leverage piecewise-stable context in reinforcement learning to help the agent adapt to environment change. 
With inferred belief context, our RL agents can quickly detect and adapt to abrupt changes in a gridwold environment and mujoco tasks with piecewise-stable context. For future work, we plan to leverage various deep learning techniques to improve SeCBAD, which includes replacing the GRU encoder by Transformer to better capture the long-term dependency in the input trajectory segment.

\bibliographystyle{apalike}
\bibliography{ref.bib}

\section*{Checklist}


\begin{enumerate}

\item For all authors...
\begin{enumerate}
  \item Do the main claims made in the abstract and introduction accurately reflect the paper's contributions and scope?
    \answerYes{}
  \item Did you describe the limitations of your work?
    \answerYes{}
  \item Did you discuss any potential negative societal impacts of your work?
    \answerYes{}
  \item Have you read the ethics review guidelines and ensured that your paper conforms to them?
    \answerYes{}
\end{enumerate}

\item If you are including theoretical results...
\begin{enumerate}
  \item Did you state the full set of assumptions of all theoretical results?
    \answerNA{}
        \item Did you include complete proofs of all theoretical results?
    \answerNA{}
\end{enumerate}

\item If you ran experiments...
\begin{enumerate}
  \item Did you include the code, data, and instructions needed to reproduce the main experimental results (either in the supplemental material or as a URL)?
    \answerYes{}
  \item Did you specify all the training details (e.g., data splits, hyperparameters, how they were chosen)?
    \answerYes{}
        \item Did you report error bars (e.g., with respect to the random seed after running experiments multiple times)?
    \answerYes{}
        \item Did you include the total amount of compute and the type of resources used (e.g., type of GPUs, internal cluster, or cloud provider)?
    \answerYes{}
\end{enumerate}

\item If you are using existing assets (e.g., code, data, models) or curating/releasing new assets...
\begin{enumerate}
  \item If your work uses existing assets, did you cite the creators?
    \answerYes{}
  \item Did you mention the license of the assets?
    \answerYes{}
  \item Did you include any new assets either in the supplemental material or as a URL?
    \answerNo{}
  \item Did you discuss whether and how consent was obtained from people whose data you're using/curating?
    \answerYes{}
  \item Did you discuss whether the data you are using/curating contains personally identifiable information or offensive content?
    \answerNo{}
\end{enumerate}

\item If you used crowdsourcing or conducted research with human subjects...
\begin{enumerate}
  \item Did you include the full text of instructions given to participants and screenshots, if applicable?
    \answerNA{}
  \item Did you describe any potential participant risks, with links to Institutional Review Board (IRB) approvals, if applicable?
    \answerNA{}
  \item Did you include the estimated hourly wage paid to participants and the total amount spent on participant compensation?
    \answerNA{}
\end{enumerate}

\end{enumerate}


\newpage
\appendix
\section{Appendix}

\subsection{Bound Derivation}
\label{Appendix:ELBO}

The variational bound for our model can be derived using importance weighting and Jensen's inequality.
We use $\tilde{c} \coloneqq c^{t-G_t+1:t}$ for simplicity:

\begin{align*}
    & \log p(\tau^X_{t-G_t:t} | a_{t-G_t:t-1}, s_{t-G_t}, G_t) \\
    &= \log \int p(\tau_{t-G_t:t}^X, \tilde{c}|a_{t-G_t:t-1}, s_{t-G_t}, G_t) \frac{q_\phi(\tilde{c}|G_t, \tau_{t-G_t:t})}{q_\phi(\tilde{c}|G_t, \tau_{t-G_t:t})} d \tilde{c} \\
    &= \log \mathbb{E}_{q_\phi} \left[ \frac{p(\tau_{t-G_t:t}^X, \tilde{c}|a_{t-G_t:t-1}, s_{t-G_t}, G_t)}{q_\phi(\tilde{c}|G_t, \tau_{t-G_t:t})} \right] \\
    &= \log \mathbb{E}_{q_\phi} \left[ \frac{p(\tau_{t-G_t:t}^X| \tilde{c},a_{t-G_t:t-1}, s_{t-G_t}, G_t)p(\tilde{c}|G_t)}{q_\phi(\tilde{c}|G_t, \tau_{t-G_t:t})} \right] \\
    &\geq \mathbb{E}_{q_\phi} \left[ \log p(\tau_{t-G_t:t}^X| \tilde{c},a_{t-G_t:t-1}, s_{t-G_t}, G_t) \right] - \mathbf{D}_{KL}(q_\phi(\tilde{c}|G_t, \tau_{t-G_t:t}) \| p(\tilde{c}|G_t)) \\
    &= \sum_{i=t-G_t+1}^{t} \mathbb{E}_{q_\phi} \left[ \log p(x_i|G_t,\tilde{c}) + \log p(s_i, r_{i-1}|G_t, \tilde{c}, s_{i-1}, a_{i-1}) \right] - \mathbf{D}_{KL} \left( q_\phi \| p(\tilde{c}|G_t) \right)
\end{align*}



\newpage
\subsection{Algorithm Framework}

In this Section, we provide the detailed training and evaluation algorithms of SeCBAD.
The training algorithm is detailed in Algorithm \ref{algo-training}. The evaluation algorithm is detailed in Algorithm \ref{algo-eval}.


\begin{algorithm}[h]
   \caption{SeCBAD Training Algorithm}
   \label{algo-training}
\begin{algorithmic}
   \State {\bfseries Input:} buffer $\mathcal{B}$, imagination horizon $H$, interacting step $T$, batch size $B$, batch length $L$, number of trajetories $N$.
   
   \State Initialize buffer $\mathcal{B}$ with $S$ random seed episodes.
   \While{ not converged } \Comment{{\color{cyan}\emph{Parameter Optimization}}}
        \For {$c=1,\dots,C$} 
            \State Draw $B$ data sequences $\{(s_t,a_t,r_t,x_t)\}_{t=k}^{k+L}$ from $\mathcal{B}$ 
            \State Calculate 
            $p(G_t|\tau_{1:t})$ using Equation (\ref{Eq:postG_norm}).
            \State Sample $g \sim p(G_t|\tau_{1:t})$.
            \State Infer belief state $q_\phi(c^{t-G_t+1:t}|G_t=g, \tau_{t-G_t:t})$.
            \State Sample $k \in \mathbb{N}$ from $[t-g+1, t]$. \Comment{{\color{cyan}\emph{Calculate the ELBO: sample 1 reconstruction step}}}         
            \State Predict observable context: $p_\theta (x_k|c, G_t=g)$
            \State Predict reward: $p_\theta(r_{k-1}|c, s_{k-1}, a_{k-1}, G_t=g)$, and state $p_\theta(s_{k}|c, s_{k-1}, a_{k-1}, G_t=g)$. 
            \State Update $\phi,\theta$ using Equation (\ref{Eq:elbo}). 
            \State Update $\psi$ using Equation (\ref{Eq:RLloss}).
        \EndFor
        \State Reset environment and get $s_1, x_1$.
        \For{$t=1,\dots,T$} \Comment{{\color{cyan}\emph{Data Collection}}}
            \State Calculate $p(G_t|\tau_{1:t})$ using Equation (\ref{Eq:postG_norm}).
            \State Calculate belief $b_t(c)$ using Equation (\ref{Eq:q-marginalize}).
            \State Compute $a_t \sim \pi_\psi(a_t|s_t, b_t)$ with action model.
            \State Add exploration noise to action.
            \State Execute $a_t$ and get $x_{t+1}, s_{t+1}, r_t$.
        \EndFor
        \State Add experience to buffer $\mathcal{B} = \mathcal{B} \cup \{(s_t,a_t,r_t, x_t)_{t=1}^T\}$
    \EndWhile
\end{algorithmic}
\end{algorithm}

\begin{algorithm}[h]
   \caption{SeCBAD Evaluation Algorithm}
   \label{algo-eval}
\begin{algorithmic}
   \State {\bfseries Input:} interacting step $T$, encoder $q_\phi$, decoder $p_\theta$, policy $\pi_\psi$.
    
    \State Reset environment and get $s_1$.
    \For{$t=1,\dots,T$}
        \State Calculate $p(G_t|\tau_{1:t})$ using Equation (\ref{Eq:postG_norm}).
        \State Choose $g = \argmax p(G_t|\tau_{1:t})$.
        \State Compute $b_t(c) = q_\phi(c^{t-G_t+1:t}|G_t=g, \tau_{t-G_t:t})$.
        \State Compute $a_t \sim \pi_\psi(a_t|s_t, b_t)$ with action model.
        \State Execute $a_t$ and get $x_{t+1}, s_{t+1}, r_t$.
    \EndFor
\end{algorithmic}
\end{algorithm}

\newpage
\subsection{Include $x$ in observation probability}
\label{Appendix:obsx}


In Section \ref{Sec:3recursive_inference}, we recursively estimate the posterior of current segment length $p(G_t|\tau_{1:t})$, which conditions the segment length on the observation $\tau_{1:t}$ as conditions to keep consistent with the case in evaluation. However, During training, we can use more information to help us get a more accurate estimation. For instance, we can include the observable contexts $x$ into $\tau_{1:t}$, like $\tau_{1:t}^X$ and update the corresponding equations as follows:

$$
    p(G_t = i, \tau_{1:t}^X) = \sum_{k=1}^{t-1} p(G_{t-1} = k, \tau_{1:t-1}^X) \cdot p(G_t = i| G_{t-1} = k) \cdot p(s_t, x_t, a_{t-1}, r_{t-1}|\tau_{t-i:t-1}^X)
$$

The difference mainly lies in the observation probability term:

\begin{align*}
    & p(s_t, x_t, a_{t-1}, r_{t-1} | \tau_{t-i:t-1}^X) \\
    &= K \mathbb{E}_{p} \left[ p(s_t, r_{t-1}|s_{t-1},a_{t-1},c^{t-i+1:t}) p(x_t|,c^{t-i+1:t}) \right]
\end{align*}

The observation probability term now estimates the data likelihood not only for the next state-reward pair, but also for the next observable context $x_t$.



There are many other choices here for how to choose the information during training. 
It is also possible to only incorporate the observable contexts $x$ instead of using $\tau_{1:t}^X$ for simplicity.
We can still use the online inference method like we stated above, or we can use some more simple and heurstic methods to derive the posterior.
Since we can directly access $x$ from the simulator, it is possible for us to calculate the posterior in advance, which can help to save a lot of computation during training time.

\subsection{Discussion with Related Works}
\label{appendix:markov-context}

Recently, some works \citep{ren2022reinforcement, feng2022factored} assume intra-episode context changes and model context $c_t$ at each time step, more specifically with Markovian transition on latent contexts.
Since Markovian assumptions are made in latent context space, these  methods are theoretically able to capture the non-Markovian pattern in observable context space.
However, as discussed in our paper, one of the key challenges  in deploying RL to real-world applications is to adapt to variations of unknown environment contexts that stay stable for a stochastic period. 
In contrast to general non-stationary environments, our problem setting assumes piece-wise stable context as special problem structure, which can be further exploited to improve performance against methods for general non-stationary environments. 
By contrast,
we propose to jointly infer the segment structure as well as the segment latent context. By doing so, we can avoid solving non-stationary environments in general case.
In our paper, we implement a one-step VariBAD to represent the methods that \li{model context $c_t$ at each time step}.
For each time step, we use the whole trajectory as input just like VariBAD. However, instead of decoding the whole trajectory, we only make the agent to decode the current time step.
%

\subsection{Bayesian Update Rule}
\label{app:bayes_update}
In Section \ref{Sec:4-1gridworld}, we use the Bayesian belief update rule to calculate the belief distribution.
Let $K$ denotes the number of grids, then $b(\theta) \in \mathbb{R}^K$ is a simplex which refers to the probability location of the goal state and therefore a categorical distribution.
As for the belief update rule, we have

$$
b'(\theta) = b(\theta|s,a,s',r) \propto b(\theta) p(r | s, a, \theta)
$$

In above equation, we omit the transition term since the transition keeps stationary across the episode. As for the reward term $p(r|s,a,\theta)$, it can be derived through the definition. 
For the $k$-th grid, suppose the observed state is the $k'$-th grid. Then, if $k = k'$, we can get the reward term as $p(r=1|\theta = k, s=k', a) = p_1$ and $p(r=0|\theta = k, s=k', a) = 1 - p_1$. 
If $k \neq k'$, we can get the reward term as $p(r=1|\theta = k, s=k', a) = p_0$ and $p(r=0|\theta = k, s=k', a) = 1 - p_0$. 
In this way, we can obtain the accurate belief without approximation.

\subsection{Hyper parameters and Implementation Details}
\label{Appendix:implementation}

\paragraph{Network Architecture}
For Half-Cheetah environments, we use a GRU\citep{cho2014learning} with 128 units as the dynamics model of the encoder $q_\phi$. For each time step, $q_\phi$ receives an encoded state, action and reward as input. The encoded state, action and reward are the output three single layer fully connected networks of respective size $[16, 32, 16]$ with ReLU as activation function. We assume the latent distribution are $5$-dimensional Gaussians with predicted mean and standard deviation. As for the transition model
$p_\theta(s_{t+1}|s_t, a_t, c)$, reward model $p_\theta(r_t|s_t, a_t, c)$ and observable context model $p_\theta(x_t|c)$, 
the output of all the decoders are Gaussian distributions with mean parameterized by fully connected network (with hidden size $[64, 32]$ and ReLU activation) and fixed identity covariance. As for the Ant-Cheetah environments, we use 10-dimensional Gaussians as the latent distribution. The other settings are kept the same as in Half-Cheetah environments.

As for the policy training part, we adopt PPO \citep{schulman2017proximal} to learn the policy $\pi_{\psi}(a_t|s_t,b_t(c))$. The actor and critic are parameterized by fully connected network of size $[128, 128]$ with Tanh as activation function.

\paragraph{Training Details}
The inputs of both actor and critic is $s_t$ and $b_t$. 
To parameterize the distribution $b_t$, we use its mean and standard deviation.
For simplicity, we only incorporate one possible segment length $g$ into consideration and this approximation will not cause obvious performance drop according to our test.
During training, we randomly sample $g$ from the posterior, while during evaluation, we use the most possible $g$.
Then, we calculate the corresponding belief $b_t$ using $g$, and use the mean and standard deviation as the inputs.
During training, we use the posterior $p(G_t|x_{1:t})$ which conditions only on the observable contexts to simplify the computation.
During evaluation, we then use the posterior $p(G_t|x_{1:t})$ like we stated in Section \ref{Sec:3recursive_inference} since we can no longer access the training only information $x$.

\paragraph{Hyper parameters}

We train PPO with Adam optimizer with learning rate 7e-4, max gradient norm 0.5, clip parameters 0.1, value loss coefficient 0.5, and entropy coefficient 0.01.
{{We use different training schedule between the VAE part (encoder \& decoder) and the policy part. We use two Adam optimizers with different learning rates. The VAE optimizer uses a learning rate of 1e-3, and the policy optimizer uses a learning rate of 7e-4.}}
For the full details of hyper-parameter settings, please refer to the code repository that we will publish soon.


\paragraph{Environment Details}
{
We give detailed descriptions of the environments we used.
}
{
\textbf{Ant Direction} environment is built upon the well-known Ant Mujoco environment. For each environment step, we set a target direction in the 2D horizontal plane. The reward function is given as $v_f - 0.2 v_v$, where $v_f$ denotes the agent velocity along the target direction, $v_v$ denotes the agent velocity that is vertical to the target direction and is always non-negative. Such reward function pushes the agent moves faster along the target direction while penalizing the agent for the velocity component that is vertical to the target direction. In experiment the target direction is chosen uniformly from $[0,2\pi]$.
}

{
For \textbf{Ant Velocity}, we set a target velocity $\boldsymbol{v_t} = (v_{t,x}, v_{t,y})$ in the 2D horizontal plane. Denote the agent velocity as $\boldsymbol{v_a}=(v_{a,x},v_{a,y})$. We project the agent velocity $\boldsymbol{v_a}$ to $\boldsymbol{v_t}$ to get the forward velocity component $v_f=\boldsymbol{v_t}\cdot \boldsymbol{v_a} / \lVert\boldsymbol{v_t}\rVert_2$. The velocity component perpendicular to the target velocity is given by $v_v=\sqrt{\lVert \boldsymbol{v_a}\rVert_2^2-v_f^2}$. If forward velocity $v_f<\lVert v_t\rVert_2$, we compute the reward as $v_f-0.3 v_v$, meaning that we expect the agent moves faster along the direction of task velocity, but not along the direction vertical to task velocity. Otherwise when $v_f\ge\lVert v_t\rVert_2$, we compute reward as $-v_f + 2\lVert\boldsymbol{v_t}\rVert_2-0.3v_v$. Observe that if ignore the term $-0.3v_v$ in the reward function, we assign the largest reward when $v_f = \lVert \boldsymbol{v_t}\rVert_2$, and reward decreases as $v_f$ deviating from $\lVert \boldsymbol{v_t}\rVert_2$, thus promoting the agent follows the target velocity. In experiment, $v_{t_x}$ and $v_{t_y}$ are independently chosen from $\text{Uniform}(-3,3)$ for each segment.
}

{
\textbf{Cheetah Direction}, \textbf{Cheetah Goal} and \textbf{Cheetah Velocity} are modified from Half Cheetah Mujoco environment. The agent can only move along $x$-axis. In the original version of Half Cheetah, the reward is given by the agent velocity along the positive $x$-axis. Denote agent velocity as $v_a$ and agent location as $x_a$. In Cheetah Direction, we set the target direction to positive $x$-axis or negative $x$-axis, and the reward is $v_a$ if the target direction is positive $x$-axis or $-v_a$ otherwise. We choose the positive and negative directions with equal probability. In Cheetah Goal, we set a target location $x_t$ on the $x$-axis in each environment step, and the reward function is $-|x_a-x_t|$, with $x_t$ sampled from $\text{Uniform}(-5,5)$. The agent location $x_a$ is also included into the observation space, which is not for the original Half Cheetah, Cheetah Dir and Cheetah Velocity. For Cheetah Velocity, we set a target velocity $v_t$ on the $x$-axis, the reward is $-|v_t-v_a|$, and $v_t$ is chosen from $\text{Uniform}(-5,5)$.
}

\newpage
\subsection{Extended Experiment Results}
\label{Appendix:expMujoco}

\subsubsection{A Case Study on Ant Direction}
\label{app:casestudy}

\begin{figure}[h]
    \centering
    \includegraphics[width=0.95\textwidth]{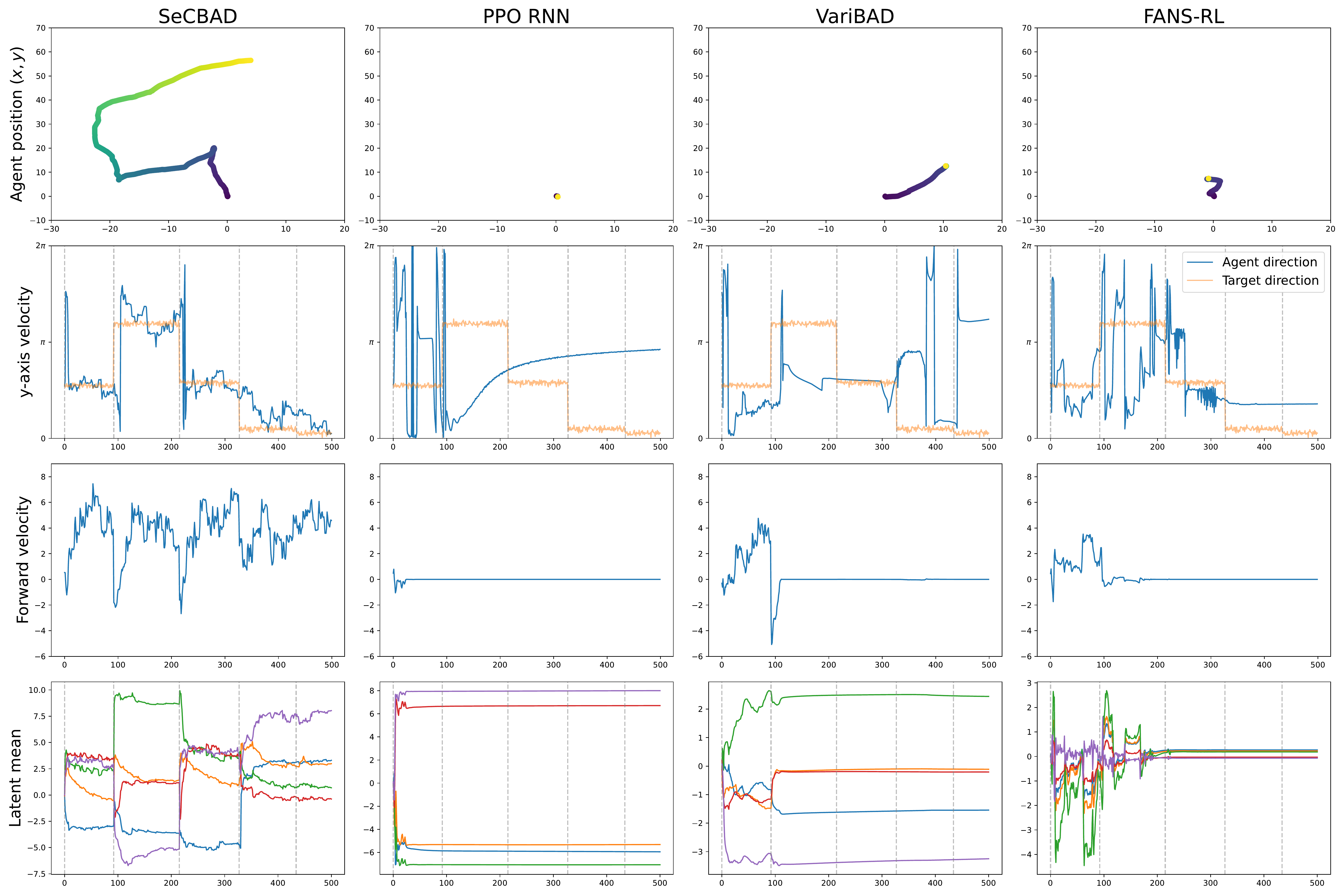}
    \caption{Behavior and the learned latent contexts of different algorithms on Ant Direction}
    \label{fig:app-behavior}
\end{figure}


{{
To better understand the outperformance of SeCBAD, we show the behavior of SeCBAD and other baselines and the learned latent contexts in Figure \ref{fig:app-behavior}.
}}

{The first row shows the agent's location through time. It can be seen that the agent of PPO RNN, VariBAD, and Factor MDP choose to stay around a fixed position shortly after starting, while SeCBAD can successfully guide the agent moving around.}

{In the second row, we show the agent's direction along with the target (task) direction. The orange curve shows the time-varying task direction, and the blue curve shows the agent's actual direction.
Since the goal is to maximize the velocity along the task direction, we plot the velocity along the task direction in the third row. 
It can be concluded that SeCBAD is able to detect and adapt to the variations rapidly, and the agent is indeed moving along the task direction, while other methods fail to detect and adapt to the non-stationary task in time. After detecting the teak direction change, SeCBAD's velocity drops and then grows up as it tries to change the direction, while the velocities of other baselines are close to zero and the learned behaviors are undesirable.}

{We provide some insights into the results. We plot the mean of the latent in the fourth row of Figure \ref{fig:app-behavior}.
VariBAD \citep{zintgraf2020varibad} tries to learn a single context for the whole episode and train the decoder by reconstructing the whole trajectory. This may make the agent learn the same contexts across the episode.
Therefore the learned latent mean is smooth and thus uninformative to the changing environment. Please refer to Appendix \ref{ablation:numberofsegment} for detailed analysis.
FANS-RL \citep{feng2022factored} assumes that the contexts evolve in a Markovian pattern. PPO-RNN \citep{hausknecht2015deep} treats the problem in a general form as a POMDP and leverages little information about the context pattern. It is hard for these two methods to learn a meaning for contexts, especially in complex environments like Ant Direction.
However, as for SeCBAD, thanks to the joint inference framework and segment decoder, the learned latent mean can change abruptly when the inferred segment structure changes, which informs the policy that the contexts have changed. The behavior in Figure \ref{fig:app-behavior} shows that the agent can learn to adapt to the variations rapidly with this accurate enough belief.
We provide more case studies on other tasks in Section \ref{Sec:4-2mujoco} in Appendix \ref{ablation:casestudy2}.
}

\subsubsection{Prior Analysis}
\label{app:prior}

\begin{figure}[h]
    \centering
    \includegraphics[width=0.5\textwidth]{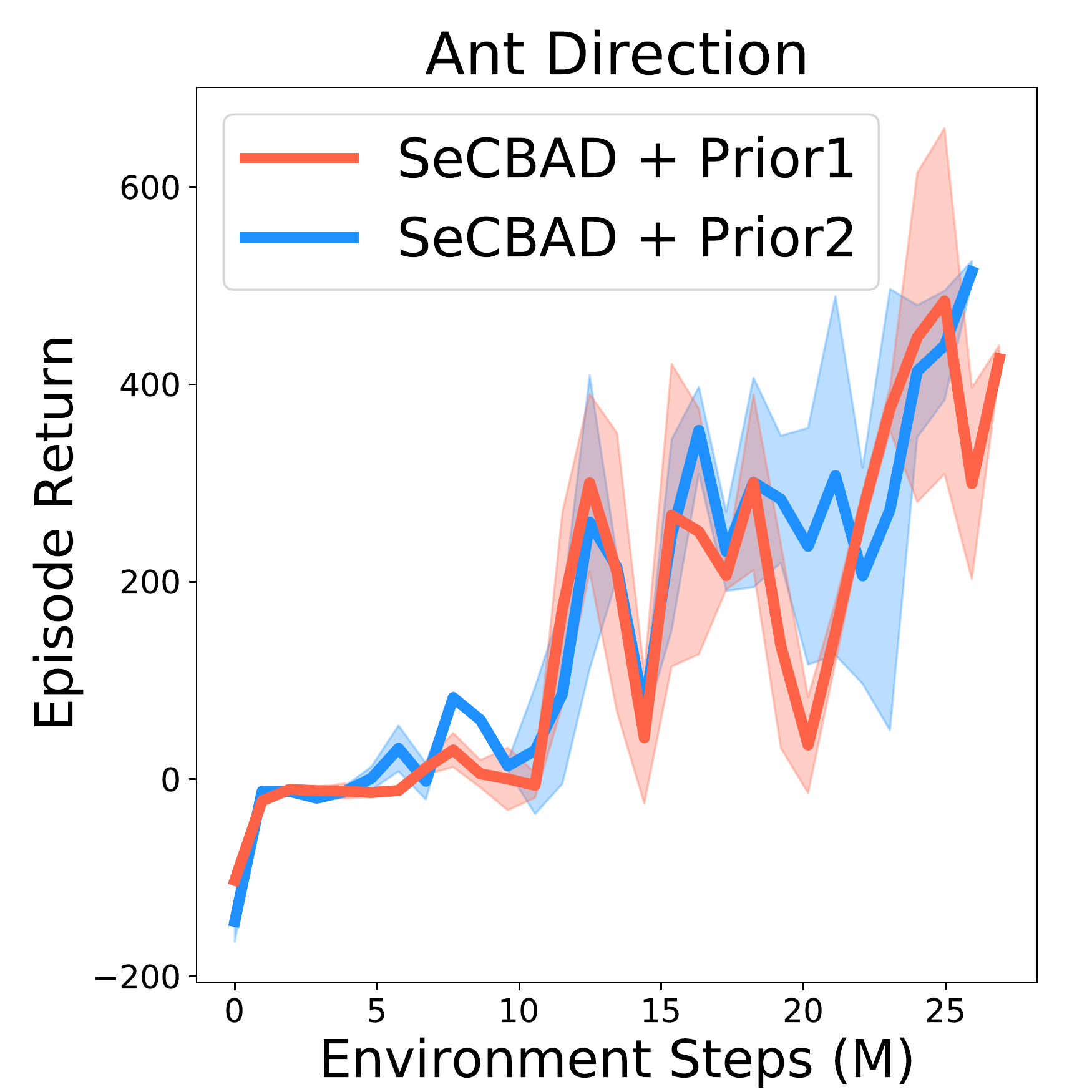}
    \caption{Analysis on the choice of prior}
    \label{fig:app-prior}
\end{figure}

In this Section, we provide the analysis on the choice of prior $p(G_t = k | G_{t-1}=i)$ on Ant Direction. The results are provided in Figure \ref{fig:app-prior}.
We choose two different prior functions. For prior $1$, we let $p(G_t = 1 | G_{t-1}=i) = \frac{1}{80}$ for any $i$ and $p(G_t = i+1 | G_{t-1}=i) = \frac{79}{80}$. For prior $2$, we use a more accurate prior by rolling out the generative process several times and approximate the $p(G_t=k|G_{t-1}=i)$ from the simulated data. 
The performance using these two priors are shown above.

\begin{figure}[h]
    \centering
    \includegraphics[width=0.6\textwidth]{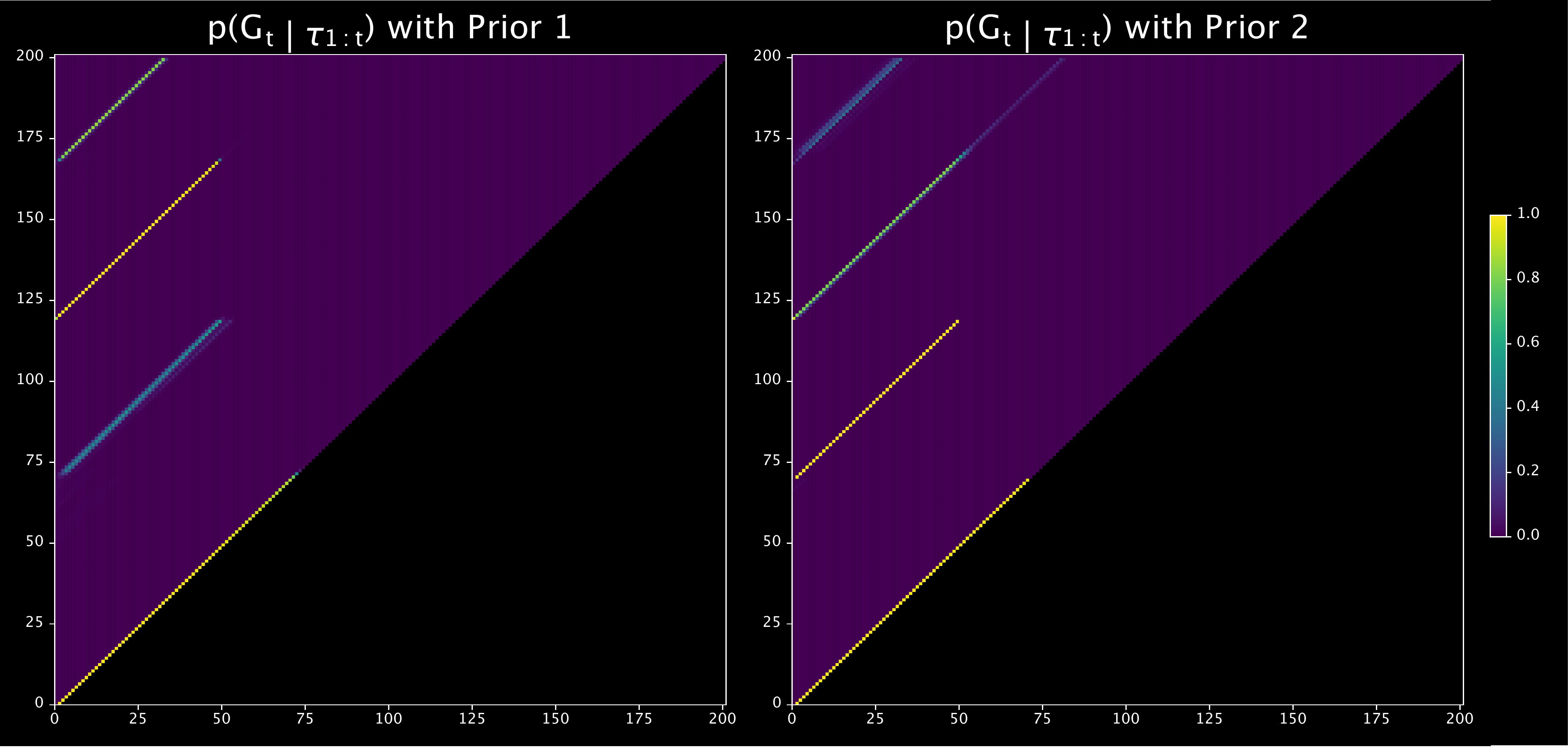}
    \caption{The segment length posterior $p(G_t|\tau_{1:t})$ calculated with two different prior functions.}
    \label{fig:app-prior-gt}
\end{figure}

It can be concluded that, two prior funciton have little performance gap. To gain more insights, we draw the probablity of $p(G_t|\tau_{1:t})$ in Figure \ref{fig:app-prior-gt}. The results show that the inferred posterior is very deterministic, and the one with approximate prior is a little bit more blurry, but still deterministic enough.
This result proves that, the observation term dominates the probability term and our method is robust with respect to the choice of prior.

\subsubsection{Ablation study on the number of segments}
\label{ablation:numberofsegment}

\begin{figure}[h]
    \centering
    \includegraphics[width=0.95\textwidth]{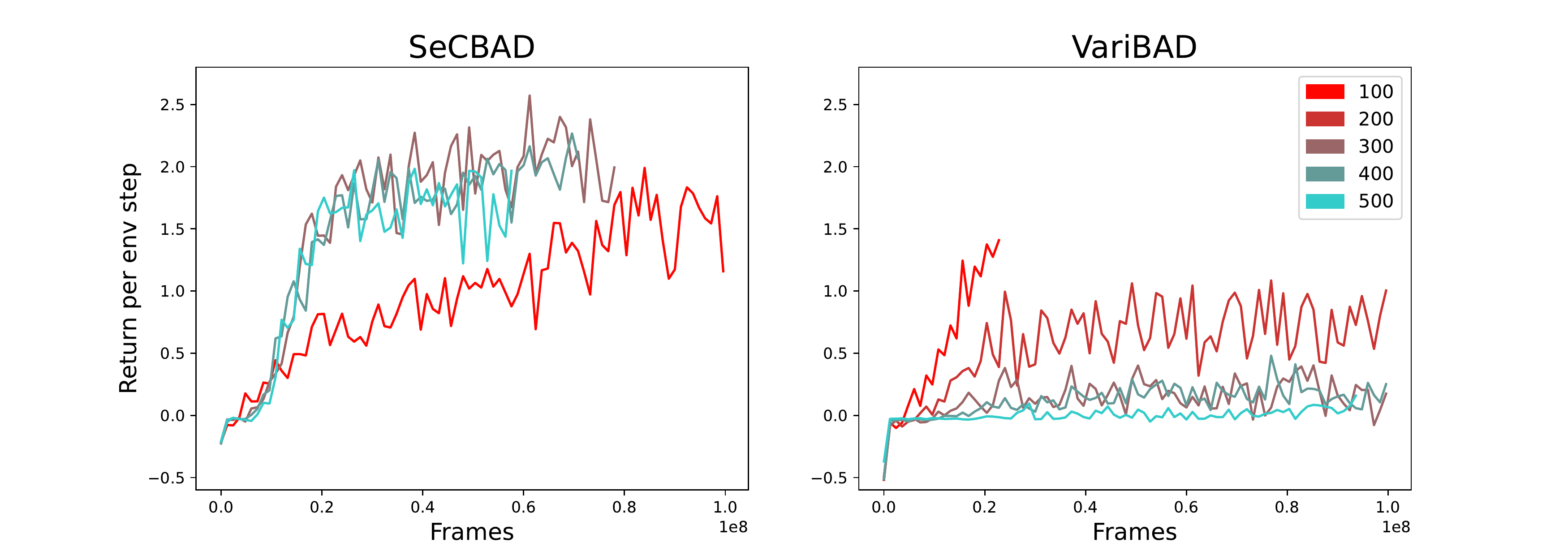}
    \caption{Ablation study on the number of segments. We keep the segment length fixed and adjust the episode length $L=\{100, 200, 300, 400, 500\}$ to adjust the number of segments.}
    \label{fig:abl-trajlen}
\end{figure}

{{
In this section, we study how the number of segments affects the performance of SeCBAD and VariBAD \citep{zintgraf2020varibad}.
As analyzed in Section \ref{Sec:4-2mujoco}, we hypothesize that methods assuming that contexts stay the same within an episode like \citep{zintgraf2020varibad} may end up learning an averaged latent contexts.
We argue that this is not related to the length of the segment but is due to the number of segments in an episode. 
During reconstruction, \citep{zintgraf2020varibad} uses the learned latent contexts at timestep $t$ to decode the transitions and rewards in the whole trajectory. Suppose there are $n$ segments in the episode, then the probability of reconstructing the correct transition and reward (i.e., those in the same segment) is only $\frac{1}{n}$.
The noisy even erroneous training signal may exacerbate as $n$ grows and end up in learning averaged latent contexts for the whole episode.
}}

{{
To look further into this hypothesis, we conduct an ablation study on the number of segments. We choose the Ant Direction environment and change the max episode length $L \in \{100, 200, 300, 400, 500\}$ while letting the average segment length be fixed. Then the expected number of segments $n$ may vary. We plot the averaged reward per step instead of the accumulated rewards to cancel out the effect of $L$ and study the effects on $n$.
}}

{{As illustrated in Figure \ref{fig:abl-trajlen}, for SeCBAD, $n$ has little effect on performance. However, for VariBAD, the performance deteriorates as the grows. The results fit the above analysis and show the significance of joint inference on the segment structure and context belief.
}}

\subsubsection{Ablation study on using $p(G_t|\tau_{1:t})$}
\label{ablation:pgt}

\begin{figure}[h]
    \centering
    \includegraphics[width=0.6\textwidth]{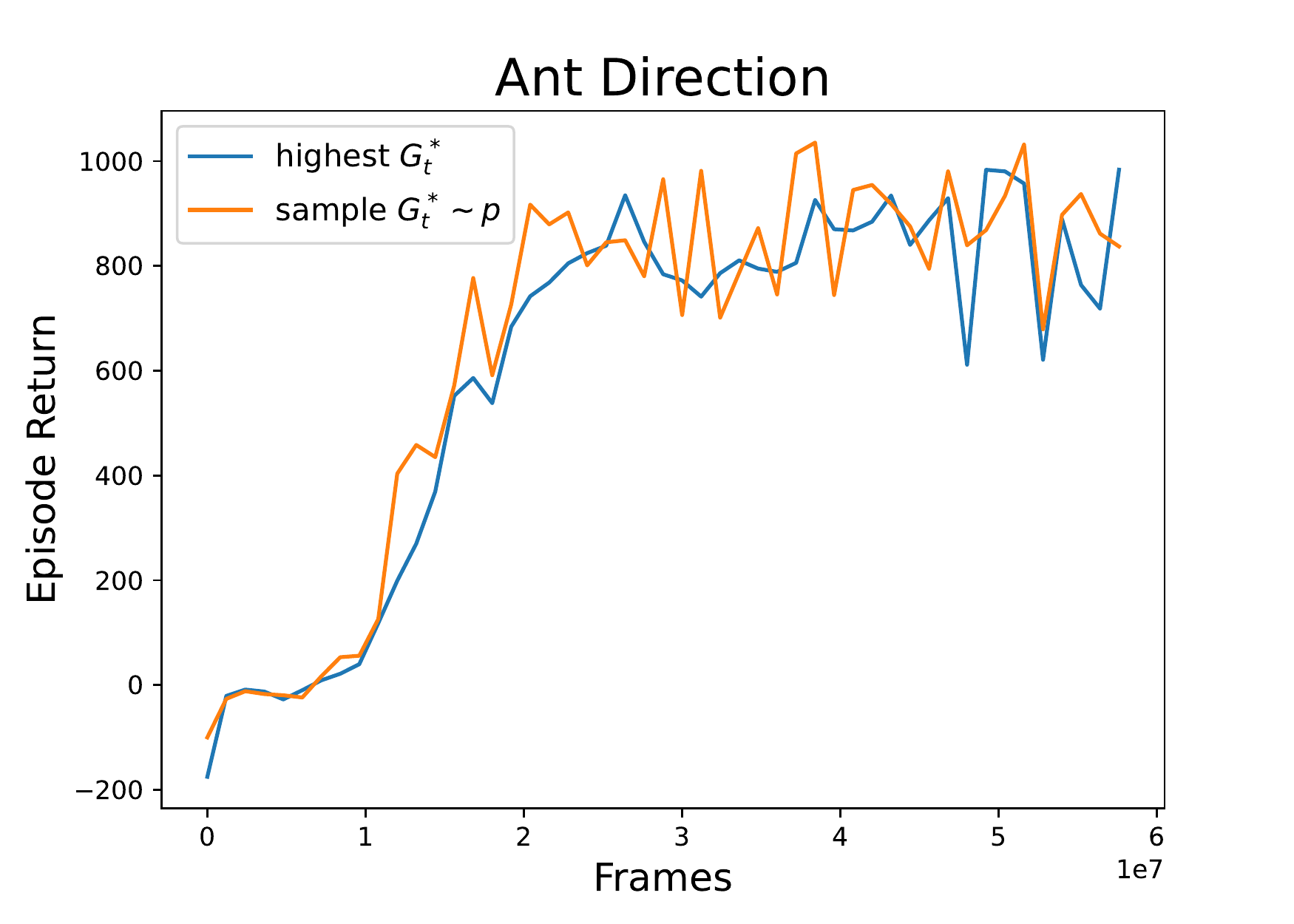}
    \caption{Ablation study on how to use the inferred $p(G_t|\tau_{1:t})$.}
    \label{fig:abl-pgt}
\end{figure}

{{
During the implementation, we may face a choice on how to use the inferred segment structure $p(G_t^*|\tau_{1:t})$. Since there are $t$ possible choices for $p(G_t|\tau_{1:t})$ in timestep $t$, it's hard to directly use the full distribution. 
One option is to use the $G_t$ with the highest probability. 
Another option is to sample $G_t \sim p$ to approximate the full distribution. 
In this section, we perform an ablation study on Ant Direction to show the performance of the two options. We keep other parameters fixed and only modify the way using $G_t^*$.}}
As shown in Figure \ref{fig:abl-pgt},
{{
the performances of the two options are very close. Therefore, for stability, we choose the $G_t^*$ with the highest probability in experiments. 
}}

\subsubsection{Ablation study on context robustness}
\label{ablation:robustness}

\begin{figure}[h]
    \centering
    \includegraphics[width=0.6\textwidth]{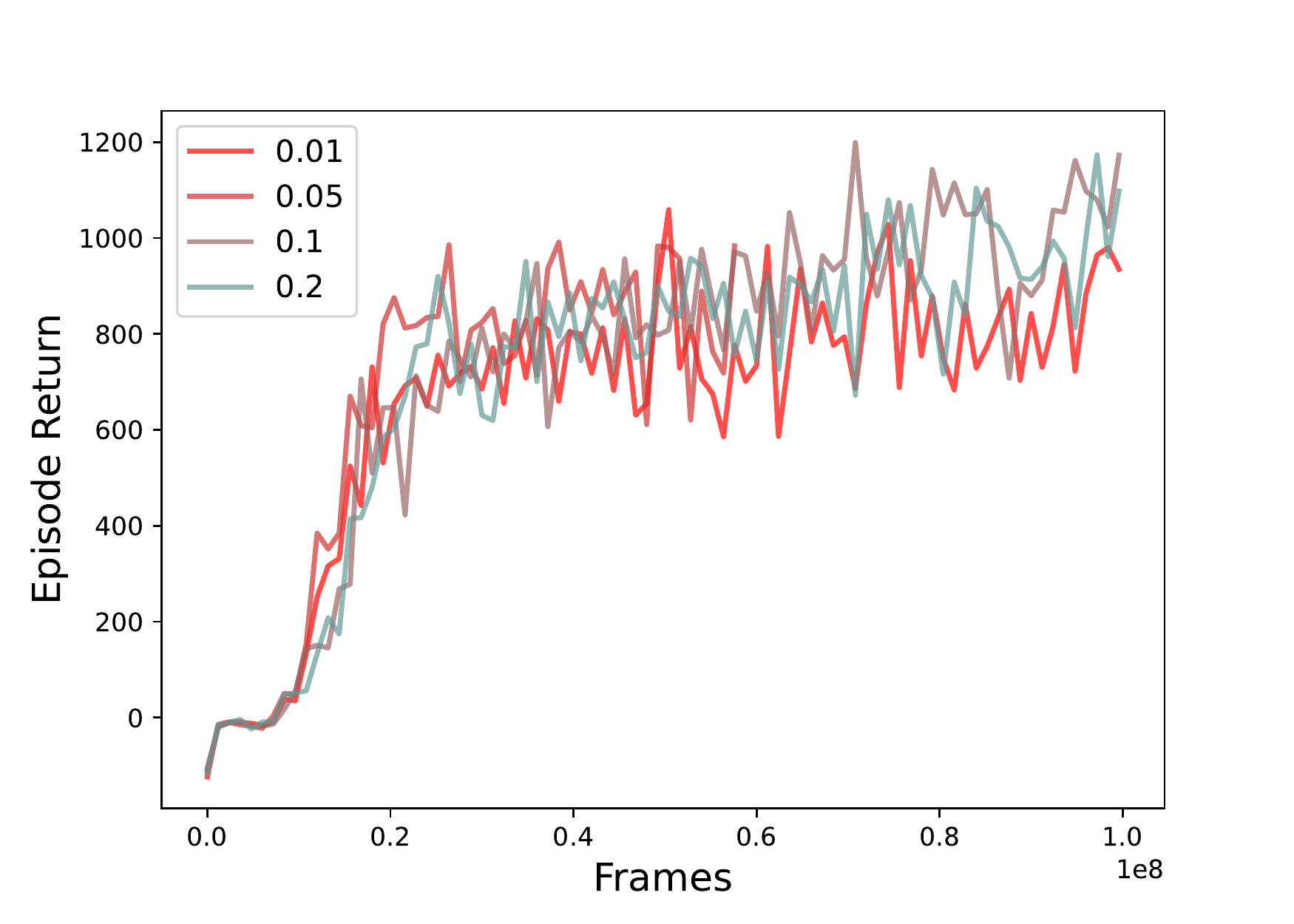}
    \caption{Ablation study on the robustness of SeCBAD to the noise within each segment. We change the standard deviation $\sigma \in \{0.01, 0.05, 0.1, 0.2\}.$}
    \label{fig:abl-robust}
\end{figure}

{{
In many real-world applications, the contexts are generally piecewise stable. However, within each segment, the contexts might be slightly noisy. Therefore, it is important to test the robustness of the proposed algorithm against the noise within each segment.
In this section, we conduct an ablation study on SeCBAD on Ant Direction by adjusting the standard deviation of the noises.
}}

{{
For each segment, we uniformly sample the mean of the contexts. Then, for each timestep within each segment, we assume the contexts follow a Gaussian distribution. We adjust the standard deviation of the Gaussian distribution to model different levels of robustness by setting $\sigma = \{0.01, 0.05, 0.1, 0.2\}$ and keep other parameters fixed.
As illustrated in Figure \ref{fig:abl-robust}, SeCBAD is able to handle noises within each segment since the performances of different $\sigma$ are very close. This further exhibit the ability of SeCBAD to solve many real-world applications.
}}

\newpage
\subsubsection{More visualizations}
\label{ablation:casestudy2}

{In this section, we provide case studies of SeCBAD and other baselines for the rest environments in Section \ref{Sec:4-2mujoco}. For the case study on Ant Direction, we refer to \ref{app:casestudy}. Given each environment, we visualize the model behavior after training for the same number of frames. We use the same random seed for SeCBAD and other three baselines, thus these algorithms are tested on environments with the same trajectory context.}

\begin{figure}[h]
    \centering
    \includegraphics[width=0.9\textwidth]{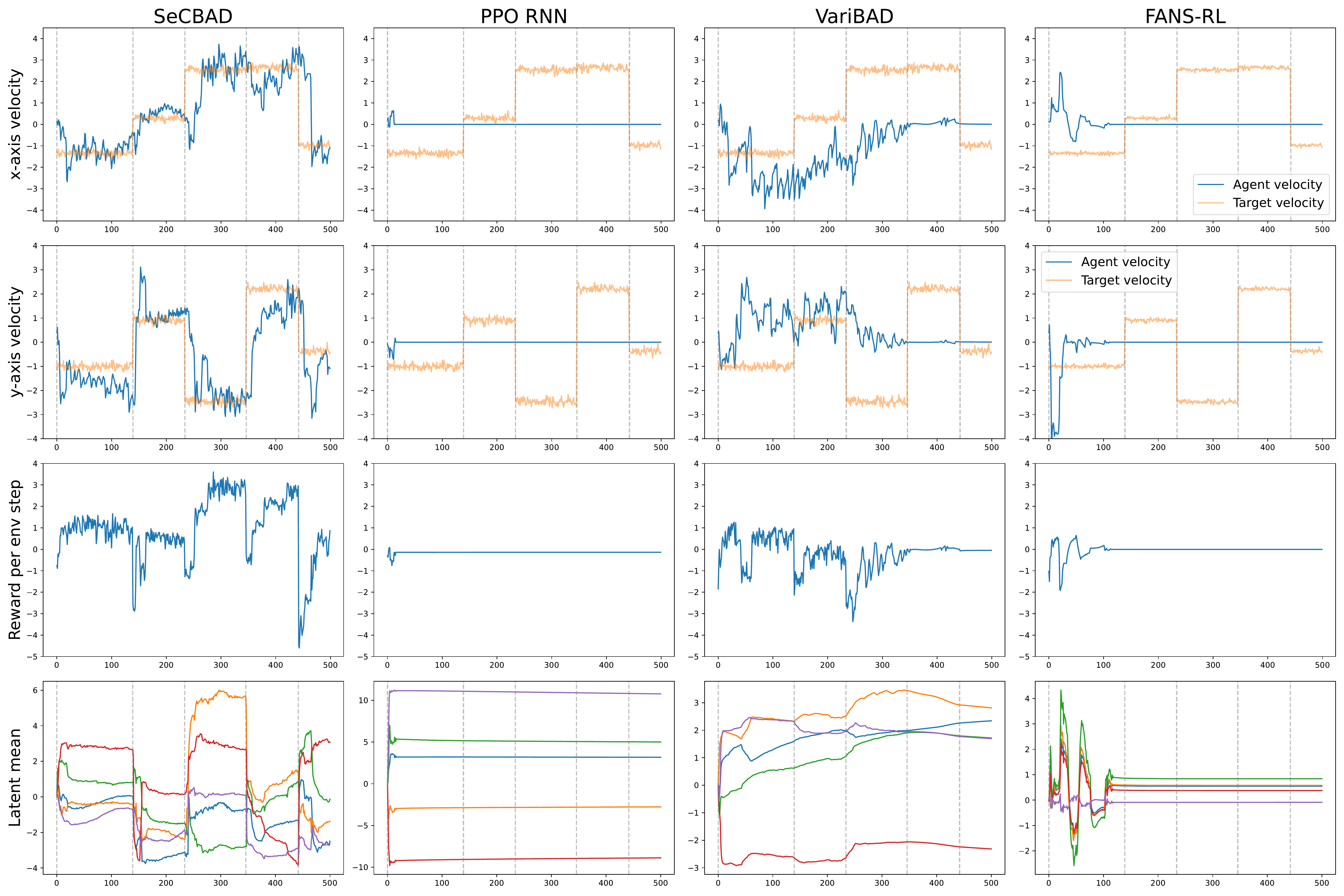}
    \caption{A case study on Ant Velocity.}
\end{figure}

\begin{figure}[h]
    \centering
    \includegraphics[width=0.9\textwidth]{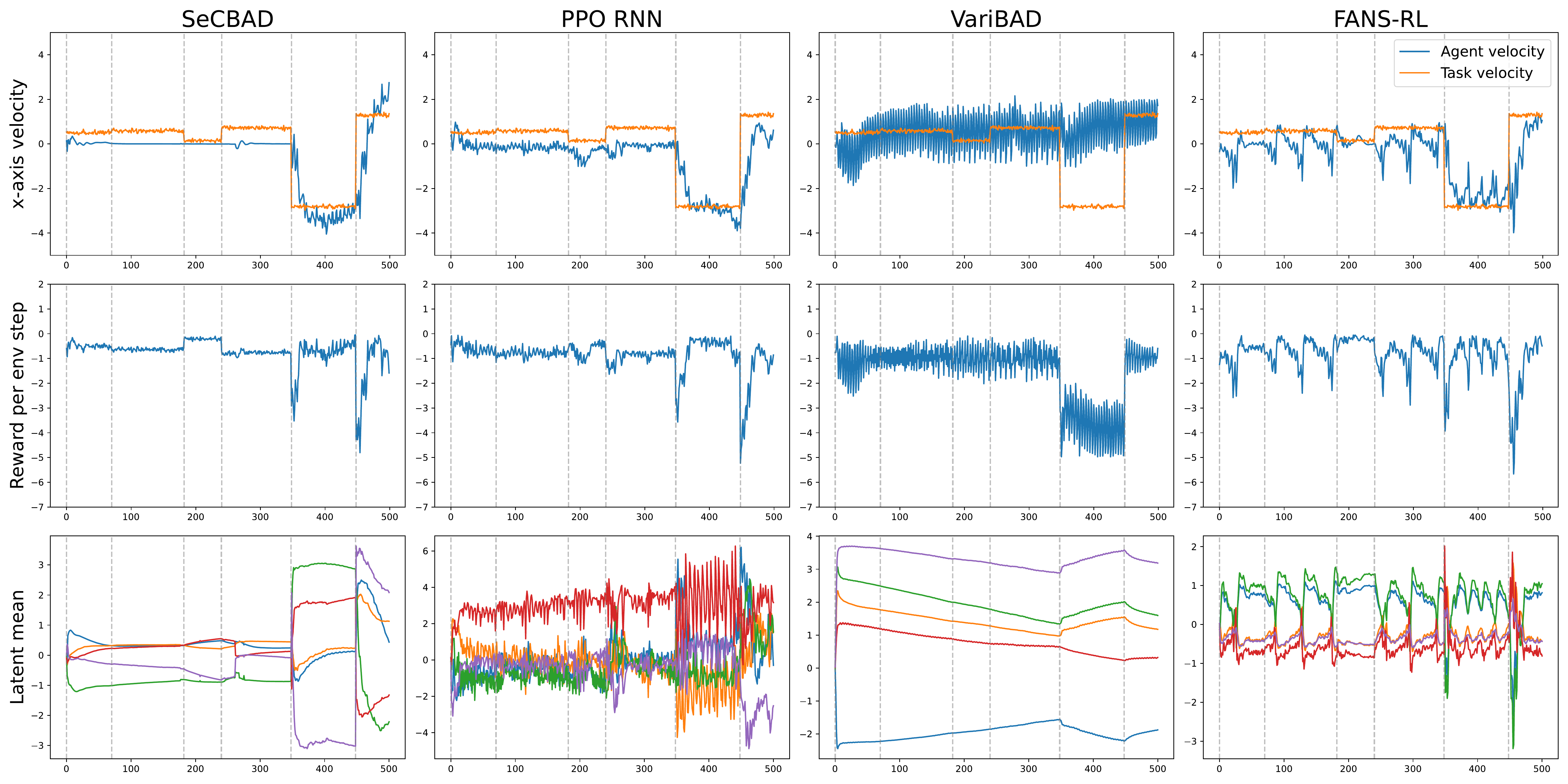}
    \caption{A case study on Half Cheetah Velocity.}
\end{figure}

\begin{figure}[h]
    \centering
    \includegraphics[width=0.9\textwidth]{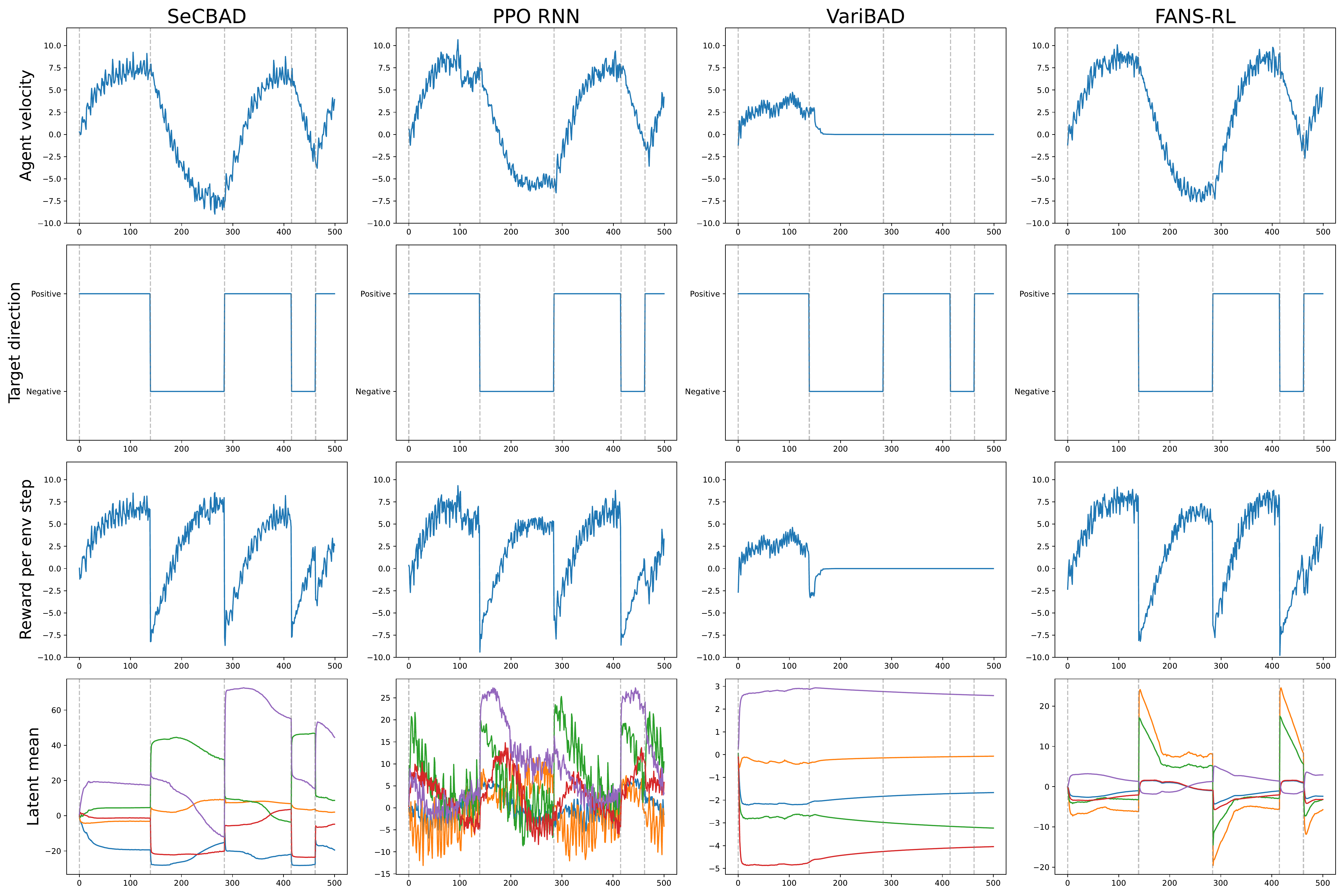}
    \caption{A case study on Half Cheetah Direction.}
\end{figure}

\begin{figure}[h]
    \centering
    \includegraphics[width=0.9\textwidth]{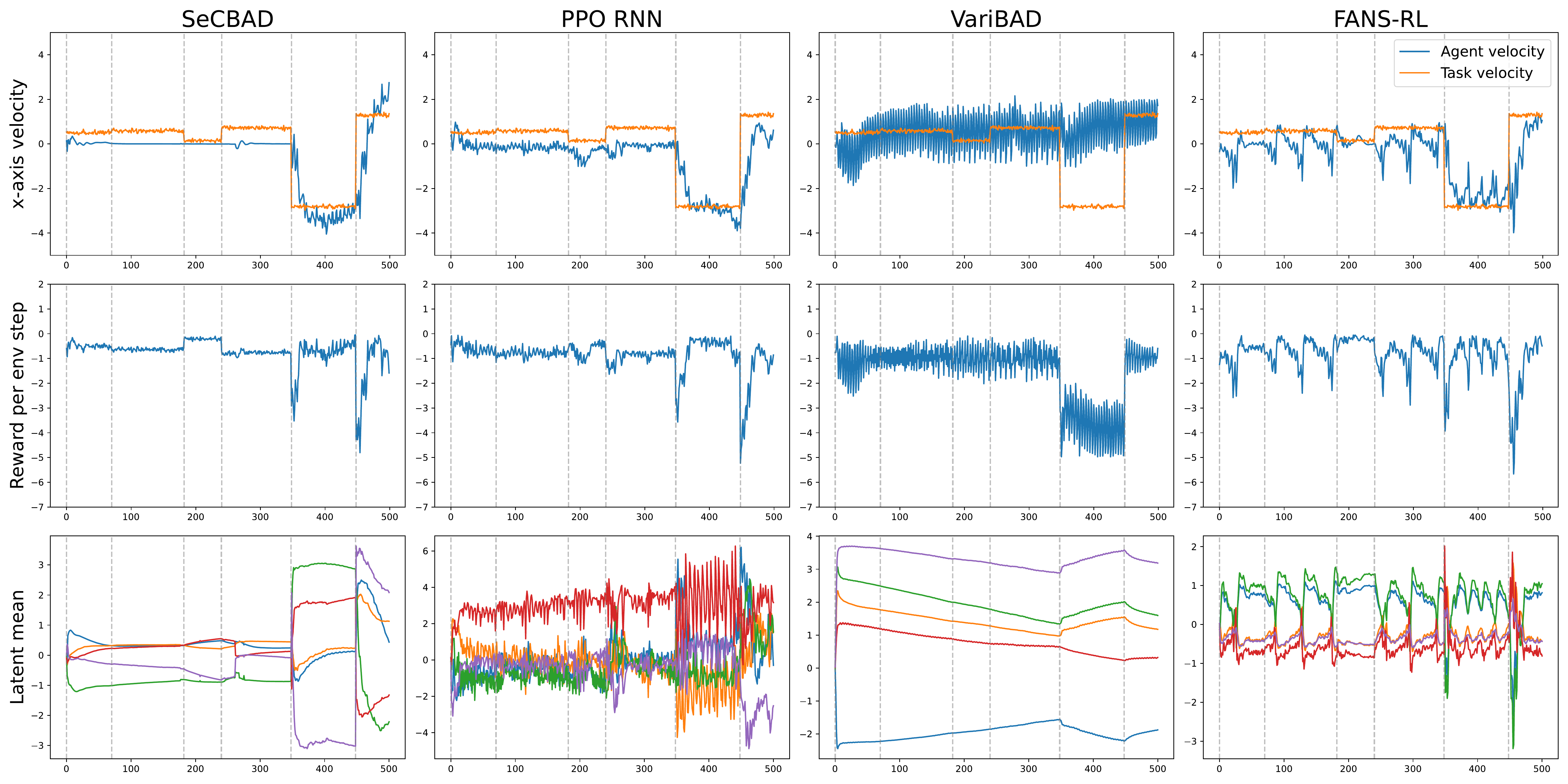}
    \caption{A case study on Half Cheetah Velocity.}
\end{figure}

\newpage
\subsection{Bandwith Control Tasks for Real Time Communication}
\label{app:rtcexp}
{{
To further illustrate the proposed LS-MDP setting can boost the deployment of RL in many real-world applications, we test SeCBAD on a real-world bandwidth control task for real-time communications (RTC) \citep{alphartc} in this section.}}

{{
RTC applications, e.g., online audio/video calls and conferences, have been greatly increasing since the global pandemic.
The most critical goal in RTC is to provide high Quality of Experience (QoE) for users, including high audio/video bitrate, low end-to-end latency, no video freeze, and few bitrate switches.
To achieve this, a bandwidth control module is needed, i.e., the RTC sender needs to decide the bitrate of outstreaming audio/video based on the network status towards the receiver.
For example, it would decrease the bitrate when observing a high end-to-end delay, otherwise, the bitrate would be increased.}}
{{
However, the ground truth network conditions are always changing in multiple items, such as available capacity, Round-Trip Time (RTT), and so on. The simple rule \textit{"decrease the bitrate when observed a high delay"} becomes unreasonable when the ground truth RTT is increased.
This reveals the system should rapidly detect and react to the network condition otherwise the users' QoE may suffer.
}}

{
To test SeCBAD for the bandwidth control problem in RTC, we use AlphaRTC \citep{alphartc} as our simulator environment.
In our reinforcement learning formulation, we use a 7-tuple of current network statistics that is visible to the agent as states $s_t$, consisting of sending rate, short-term and long-term receiving rate, loss, and delay.
The action $a_t$ is the estimated bandwidth. The reward function is formulated as  $2R / C - (D - RTT/2) - L - 1 $, where $R$ is the receiving rate in the time step, $D$ is the average delay in the time step, $L$ is the packet loss rate, $C$ and $RTT$ are the ground truth capacity and average RTT of the network.
The latent context $c_t$ here refers to the fluctuated network condition, in this section, we consider $x_t$ as the ground truth bandwidth capacity $C$, and the RTT.
The agents are trained in AlphaRTC \citep{alphartc} that simulates real-time communication processes by specifying $C$ and $RTT$. All the network statistics are normalized to the $(0, 1)$ range.}

\begin{figure}[h]
    \centering
    \includegraphics[width=0.8\textwidth]{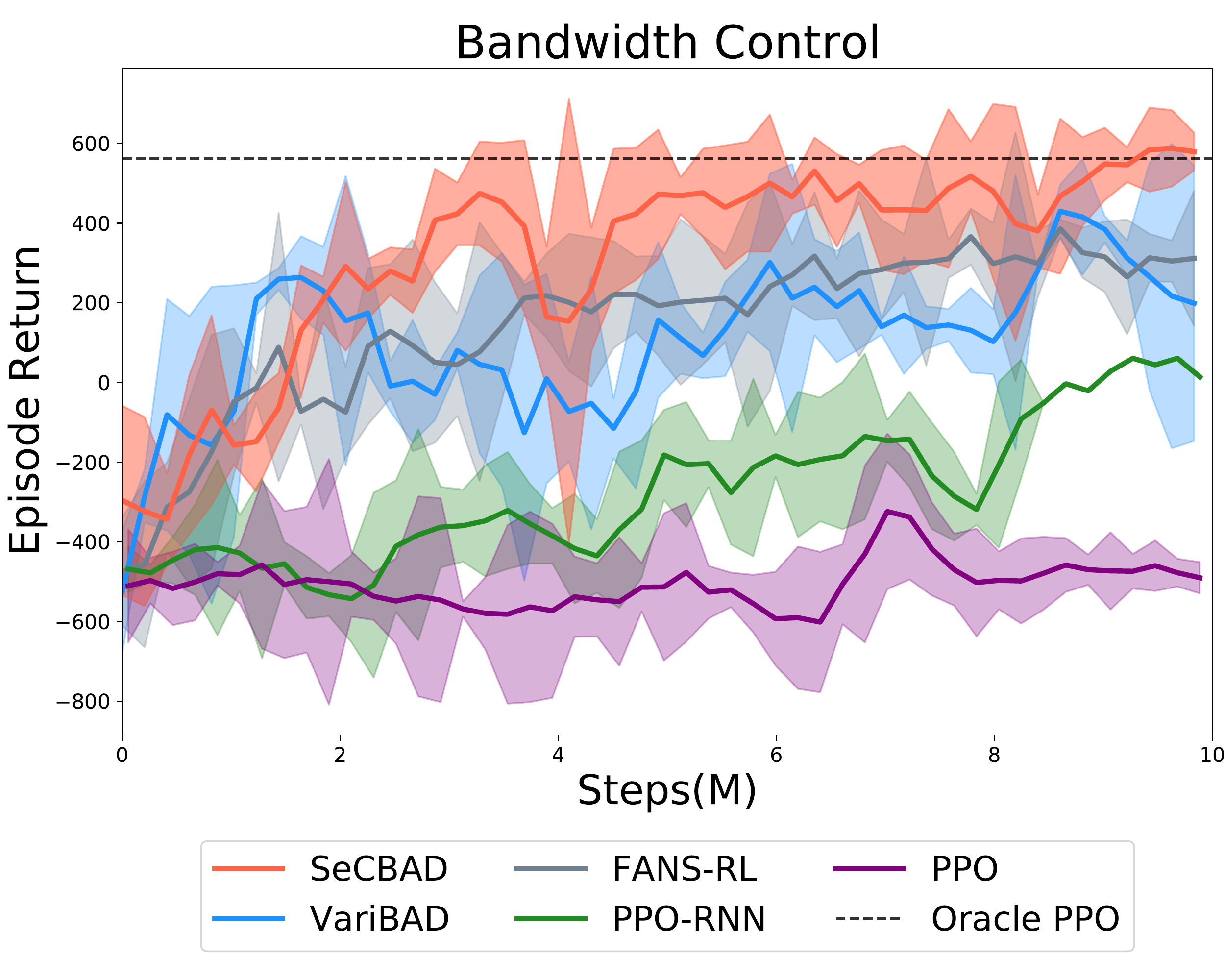}
    \caption{Experiment results on bandwidth control for RTC.}
    \label{fig:exp-rtc}
\end{figure}

{
In this experiment, we compare our method with VariBAD \citep{zintgraf2020varibad}, FANS-RL \citep{feng2022factored}, vanilla PPO \citep{schulman2017ppo} and PPO-RNN \citep{hausknecht2015deep}. 
We also incorporate oracle PPO scores by incorporating the unobservable contexts into the observable states}. {{All the methods are trained for 10 million steps and the shaded area is across 3 random seeds.}}

{As illustrated in Figure \ref{fig:exp-rtc}, SeCBAD achieves better performance than other baselines and is very close to the oracle PPO baseline score. The performances of VariBAD \citep{zintgraf2020varibad} and FANS-RL \citep{feng2022factored} are better than PPO-RNN \citep{hausknecht2015deep}, but SeCBAD outperforms both of these methods. The results suggest that SeCBAD is able to detect and adapt to the varying contexts more rapidly, which allows the policy to precisely control. This indicates the importance of the joint inference structure. 
}

\begin{figure}[h]
    \centering
    \includegraphics[width=0.8\textwidth]{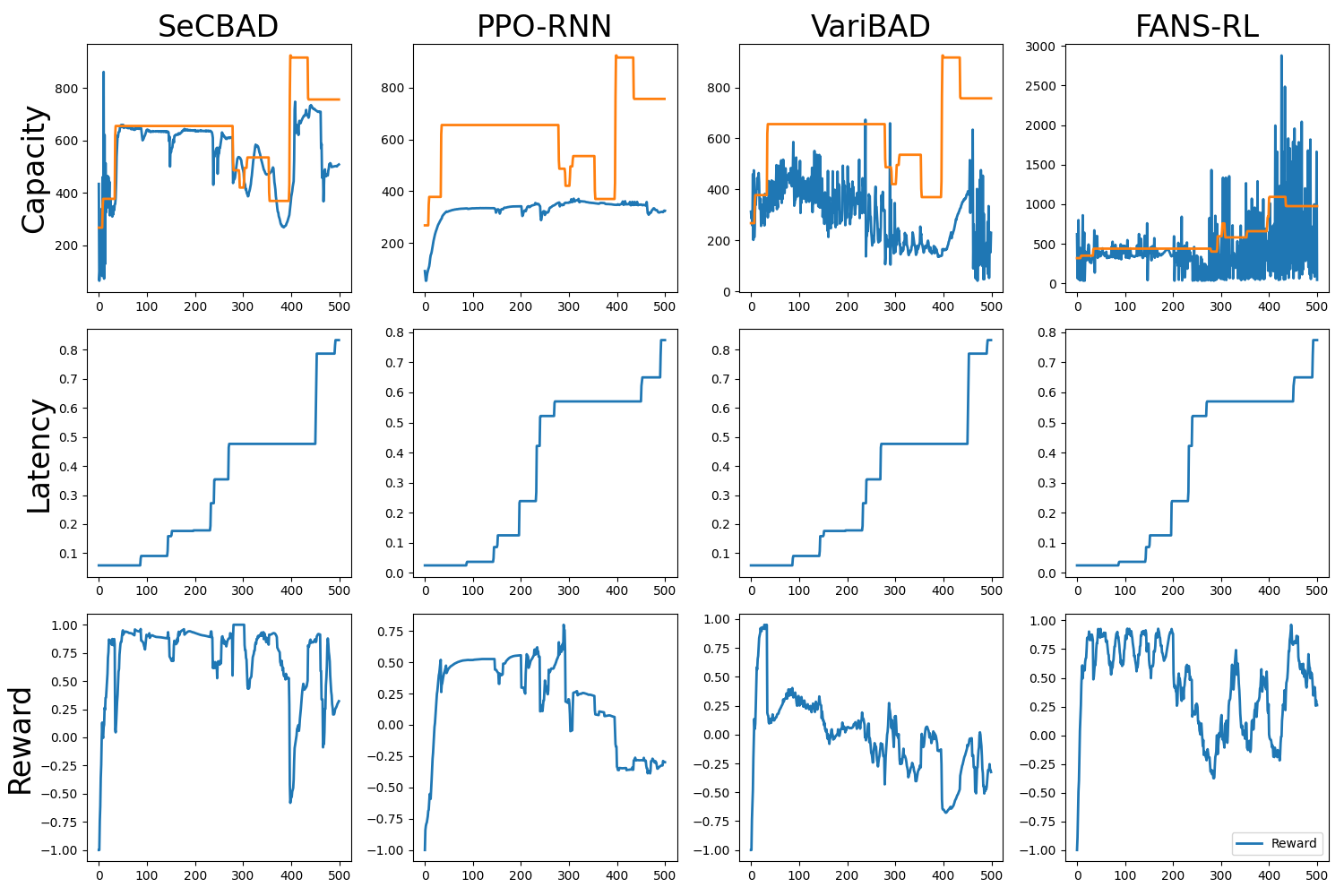}
    \caption{Case study on bandwidth control for RTC.}
    \label{fig:case-rtc}
\end{figure}

{{
Figure \ref{fig:case-rtc} shows the detailed behavior of different methods under one representative case.
}}

{{
The first row shows the agent's action (in blue) against the true available bandwidth (in orange). The second row shows the latency observed by the agent, which partially represents the mixed effect of another context, RTT. It can be observed that for SeCBAD, the agent is able to detect the change in bandwidth and adapt to the changes in time and produce a stable policy within each segment. The drops in actions are timed to coincide with the observed increasing latency. After realizing that the actual capacity is not changed, the agent can then increase the actions back to the optimal value.
However, for other baselines, the actions oscillate a lot, especially for FANS-RL \citep{feng2022factored} which is not practical since this may lead to frequent bitrate switches. VariBAD \citep{zintgraf2020varibad} and PPO-RNN \citep{hausknecht2015deep} learn a quite smooth and conservative policy that leaves a large margin between the action and the actual capacity, which may lead to a waste in the network capacity.
}}

\end{document}